\documentclass[10pt,twocolumn,letterpaper]{article}

\usepackage{cvpr}              % To produce the CAMERA-READY version
\usepackage[dvipsnames]{xcolor}

\definecolor{cvprblue}{rgb}{0.21,0.49,0.74}
\usepackage[pagebackref,breaklinks,colorlinks,citecolor=cvprblue]{hyperref}
\usepackage{multirow}
\usepackage{colortbl}
\usepackage{pifont}

\title{MIGC: Multi-Instance Generation Controller for Text-to-Image Synthesis}

\author{Dewei Zhou$^{1}$$^*$ \qquad
    You Li$^{1}$$^*$ \qquad
    Fan Ma$^{1}$ \quad
    Xiaoting Zhang$^{2}$ \quad
    Yi Yang$^{1}$$^\dag$ \\
    $^{1}$ReLER, CCAI, Zhejiang University, Zhejiang, China \quad
    $^{2}$Huawei Technologies Ltd., China \\
    {\small $^*$ Equal contribution $^\dag$ Corresponding author} \\
    {\tt\small {\{zdw1999, uli2000, mafan, yangyics\}}@zju.edu.cn, zhangxiaoting11@huawei.com} \\
    \url{https://migcproject.github.io/}
}

\begin{document}
\maketitle

\begin{abstract}

We present a Multi-Instance Generation (MIG) task, simultaneously generating multiple instances with diverse controls in one image. 
Given a set of predefined coordinates and their corresponding descriptions, the task is to ensure that generated instances are accurately at the designated locations and that all instances’ attributes adhere to their corresponding description.
This broadens the scope of current research on Single-instance generation, elevating it to a more versatile and practical dimension.
Inspired by the idea of divide and conquer, we introduce an innovative approach named Multi-Instance Generation Controller (MIGC) to address the challenges of the MIG task.
Initially, we break down the MIG task into several subtasks, each involving the shading of a single instance.
To ensure precise shading for each instance, we introduce an instance enhancement attention mechanism.
Lastly, we aggregate all the shaded instances to provide the necessary information for accurately generating multiple instances in stable diffusion (SD).
To evaluate how well generation models perform on the MIG task, we provide a COCO-MIG benchmark along with an evaluation pipeline. 
Extensive experiments were conducted on the proposed COCO-MIG benchmark, as well as on various commonly used benchmarks. The evaluation results illustrate the exceptional control capabilities of our model in terms of quantity, position, attribute, and interaction. Code and demos will be released at \url{https://migcproject.github.io/}.
\end{abstract}    

\section{Introduction}
\label{sec:intro}

\begin{figure*}[tb]
    % \vspace{-2mm}
    \setlength{\abovecaptionskip}{0.cm}
    \setlength{\belowcaptionskip}{0.cm}
    \begin{center}
        \includegraphics[width=1.0\textwidth]{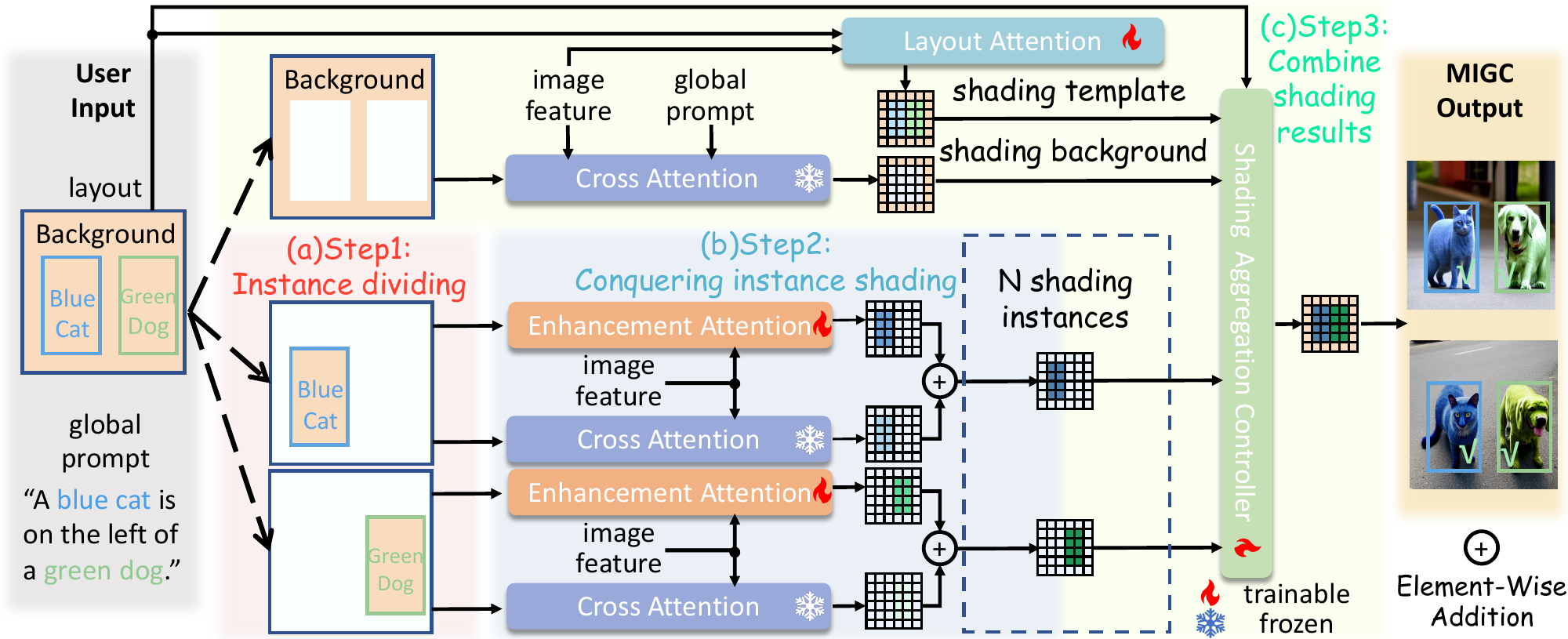}
    % \vspace{-1mm}
    \end{center}
   \caption{\textbf{Overview of our MIGC}. Stable diffusion's UNet inputs text description and image features into the Cross-Attention layer to obtain the residual feature and then adds it to the image features to determine generated content, which is like a shading process (i.e., coloring with parallel pencil lines or a block of color). In this view, MIG can be considered multi-instance shading on image features, and MIGC comprises three steps: (a) Divide MIG into single-instance shading subtasks. (b) Conquer single-instance shading with Enhancement Attention. (c) Combine shading results through Layout Attention and Shading Aggregation Controller.
} 
  \vspace{-1mm}
    \label{fig:fig_2}
  \end{figure*}
Stable diffusion~\cite{stablediffusion} has exhibited extraordinary capabilities in wild scenarios, including photography, painting, and other area~\cite{controlnet,diffusionrig,123}.
The current research mainly focuses on Single-Instance Generation, where the generated content is only required to align with the single description, including image editing, personalized image generation, 3D generation~\cite{dragdiffusion,dreambooth,anydoor,dragondiffusion,selfdiffusion,instantbooth,pfbdiff,imagic,holodiff,123}, etc. However, more practical cases where multiple instances are simultaneously generated in one image with diverse controls have been rarely explored. 
In this research, we delve into a more general task, termed as Multi-Instance Generation (MIG), incorporating all factors such as quantity, position, attribute, and interaction control into one-time generation.

\textbf{Challenges in MIG.} MIG not only requires the instance to comply with the user-given description and layout but also ensures global alignment among all instances. Incorporating this information directly into the stable diffusion~\cite{stablediffusion} often leads to failure. On the one hand, the current text encoder, like CLIP~\cite{CLIP}, struggles to differentiate each singular attribute from prompts containing multiple attributes~\cite{structurediff}.
On the other hand, Cross-Attention~\cite{attention} layers in stable diffusion lack the ability to control position~\cite{tflcg,initialimageedit,gligen}, resulting in difficulties when generating multiple instances within a specified region.

Motivated by the divide and conquer strategy, we propose the \textbf{M}ulti-\textbf{I}nstance \textbf{G}eneration \textbf{C}ontroller (MIGC) approach. This approach aims to decompose MIG into multiple subtasks and then combines the results of those subtasks. 
Although the direct application of stable diffusion in MIG is still a challenge, the outstanding capacity of stable diffusion in Single-Instance Generation could facilitate this task. 
Illustrated in Fig.~\ref{fig:fig_2}, MIGC comprises three steps: 1) Divide: MIGC decomposes MIG into multiple instance-shading subtasks \textbf{only in the Cross-Attention layers of SD} to speed up the resolution of each subtask and make the generated images more harmonious. 2) Conquer: MIGC employs an Enhancement Attention Layer to enhance the shading results obtained through the frozen Cross-Attention, ensuring successful shading for each instance. 3) Combine: MIGC obtains the shading template through a Layout Attention layer and then inputs it, together with the shading background and shading instances, into a Shading Aggregation Controller to obtain the final shading result.

\textbf{Benchmark for MIG.} To evaluate how well generation models perform on the MIG task, we propose a COCO-MIG benchmark based on the COCO dataset \cite{coco}, and this benchmark requires generation models to achieve strong control on position, attribute, and quantity \textbf{simultaneously}.

We conducted comprehensive experiments on the proposed COCO-MIG and the widely recognized COCO~\cite{coco} and DrawBench~\cite{imagen} benchmarks. When applied to the COCO-MIG benchmark, our method substantially enhanced the Instance Success Rate, increasing it from \textbf{32.39}\% to \textbf{58.43}\%. Transitioning to the COCO benchmark, our approach exhibited noteworthy improvements in Average Precision (AP), elevating it from \textbf{40.68/68.26/42.85} to \textbf{54.69/84.17/61.71}. Similarly, on DrawBench, our method demonstrated advancements across position, attribute, and count, particularly elevating the attribute success rate from \textbf{48.20}\% to \textbf{97.50}\%. Moreover, MIGC maintains an inference speed close to the original stable diffusion.

\begin{figure*}[tb]
    % \vspace{-2mm}
    \setlength{\abovecaptionskip}{0.cm}
    \setlength{\belowcaptionskip}{0.cm}
    \begin{center}
        \includegraphics[width=1.0\textwidth]{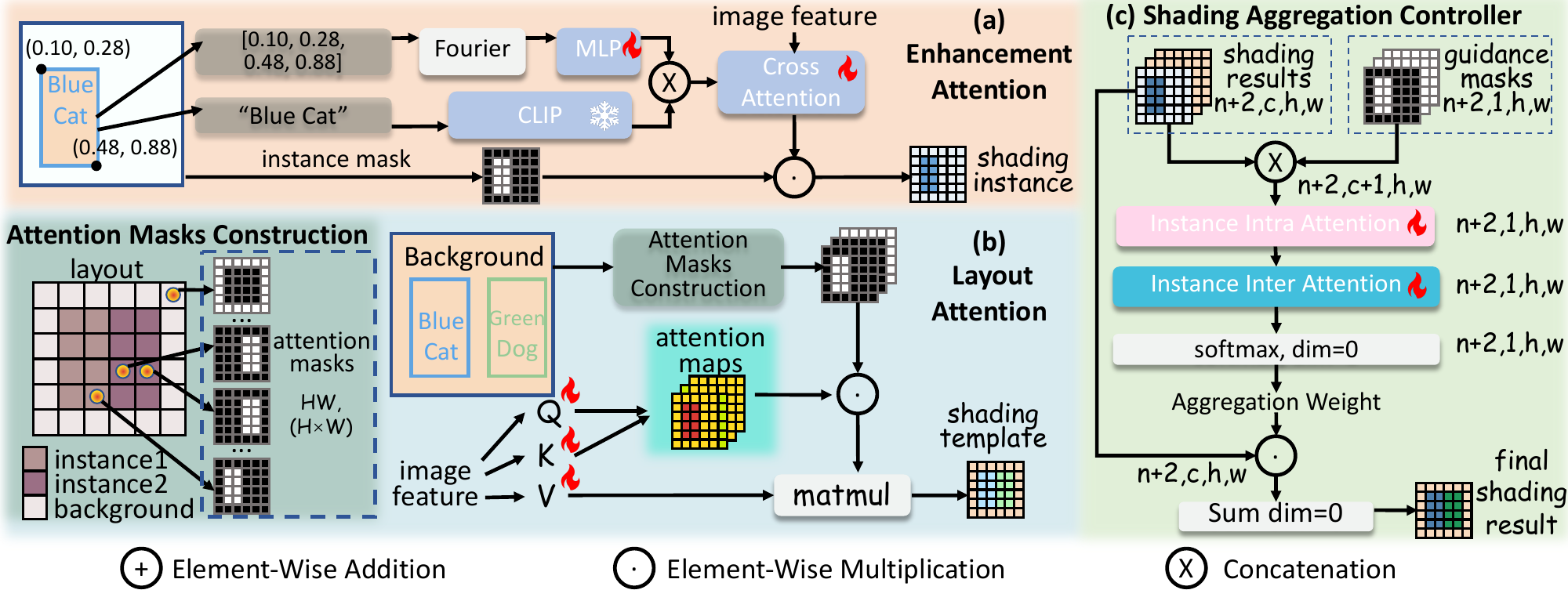}
    % \vspace{-1mm}
    \end{center}
    
   \caption{\textbf{Three main modules in MIGC}. (a) Architecture of Enhancement Attention Layer. (b) Architecture of Layout Attention Layer. (c) Architecture of Shading Aggregation Controller.} 
%   \vspace{-2mm}
    \label{fig:fig_overview}
  \end{figure*}

Our contributions are summarized as follows:
\begin{itemize}
  \item [1)] 
  To advance the development of vision generation, we present the MIG task to address prevailing challenges in both academic and industrial domains. Meanwhile, we propose the COCO-MIG benchmark to evaluate the inherent MIG capabilities of generative models.
  \item [2)]
  Inspired by the principle of divide and conquer, we introduce a novel MIGC approach that enhances pre-trained stable diffusion with improved MIG capabilities.
  \item [3)]
  We conducted extensive experiments on three benchmarks, indicating that our MIGC significantly surpassed the previous SOTA method while ensuring the inference speed was close to the original stable diffusion.
\end{itemize}

\section{Related work}
\label{sec:formatting}

\subsection{Text-to-Image Generation}
Text-to-image (T2I) Generation aims to generate high-quality images based on text descriptions. Conditional GANs \cite{gan,attngan,crossmodal,stackgan} were initially used for T2I Generation, while diffusion models \cite{glide,stablediffusion,imagen,reimagen,eDiff-I,DALL-E2,Ho2022ClassifierFreeDG,pydiff,texface} and autoregressive models \cite{MuseTG,cogview,parti} gradually replaced GANs as the foundational generator due to their more stable training and higher image quality. 

\subsection{Layout-to-Image Generation}

As text cannot precisely control the position of generated instances. Some Layout-to-Image methods~\cite{glide,tflcg,boxdiff,layoutdiff,initialimageedit} extend the pre-trained T2I model~\cite{stablediffusion} to integrate layout information into the generation and achieve control of instances' position. However, they struggle to isolate the attributes of multiple instances, thus generating images with mixed attributes. This paper proposes a novel  MIGC approach to achieve precise position-and-attribute control.

\section{Method}
\label{sec:formatting}

\newcommand{\zdwadd}{\textcolor{red}}

\subsection{Preliminaries} 
\label{preliminary}

\noindent \textbf{Stable diffusion}~\cite{stablediffusion} is one of the most popular T2I models, and it uses the CLIP~\cite{CLIP} text encoder to project texts into sequence embedding and integrate textual conditions into the generation process via Cross-Attention~\cite{attention} layers.

\noindent \textbf{Attention layers}~\cite{attention} play key roles in the interaction between various modal features. Omitting the reshape operation, the attention layer on 2D space can be expressed as:
\vspace{-0.2em}
\begin{equation}
\vspace{-0.2em}
  \mathbf{R} = \text{Softmax}(\frac{\mathbf{Q} \mathbf{K}^T}{\sqrt{d}}) \mathbf{V},\mathbf{R} \in{\mathbb{R}^{(H, W, C)}}
  \label{eq:eq_QKV}
\end{equation}
where $\mathbf{R}$ represents the output residual, and $\mathbf{Q},\mathbf{K},\mathbf{V}$ separately represents the Query, Key, and Value in attention layers, which are projected by linear layers.
% $f_Q, f_K, f_V$.

\subsection{Overview}

% 需不需要在详细介绍 满足什么要求

\noindent \textbf{Problem Definition.}  
In the Multi-Instance Generation (MIG), users will give generation models the global prompt $\mathcal{P}$, instance layout bounding boxes $\mathbb{B} = \{ \mathbf{b}^1,...,\mathbf{b}^N \}$, where $\mathbf{b}^{i} = [x_1^i,y_1^i,x_2^i,y_2^i]$, and corresponding descriptions $\mathbb{D} = \{ \mathbf{d}^1,...,\mathbf{d}^N \}$. According to user-provided inputs, the model needs to generate an image $\mathcal{I}$, in which the instance within the box $\mathbf{b}^{i}$ should adhere to the instance description $\mathbf{d}^{i}$, and global alignment is ensured in all instances.

\noindent
\textbf{Difficulties in MIG.} When dealing with Multi-Instance prompts, stable diffusion struggles with attribute leakage, i.e., 1) \textit{Textual Leakage}. Due to the causal attention masks used in the CLIP encoder, the latter instance tokens may exhibit semantic confusion~\cite{structurediff}. 2) \textit{Spatial Leakage}. The Cross-Attention lacks precise position control~\cite{tflcg}, and instances will affect the generation of each others' region.

\noindent \textbf{Motivation.} Divide and conquer is an ancient but wise idea. It first \textit{divides} a complex task into several simpler subtasks, then \textit{conquers} these subtasks respectively, and finally obtains the solution to the original task by \textit{combining} the solutions of the subtasks. This idea is highly applicable to MIG. For example, MIG is a complex task for most T2I models, while Single-Instace Generation is a simpler subtask that T2I models can solve well~\cite{dreambooth,anydoor,paint,dragondiffusion,dragdiffusion}. Based on this idea, we proposed our MIGC, which extends the stable diffusion with stronger MIG ability. We will introduce the technology details by telling ``how to divide," ``how to conquer," and ``how to combine."

\subsection{Divide MIG into Instance Shading Subtasks}

\noindent \textbf{Instance shading subtasks in Cross-Attention space.} 
Cross-Attention is the only way for text and image features to interact in stable diffusion, and the output determines the generated content, which looks like a shading operation on image features. In this view, the MIG task can be defined as performing correct Multi-Instance shading on image features, and $subtask_{i}$ can be defined as finding a single instance shading result $\mathbf{R}^{i}$ to satisfy the following:
\begin{equation}
  \mathbf{R}^{i}=\arg \min _{\mathbf{R}^{i}}(\left\|\mathbf{R}^{i}-\mathbf{R}^{correct}\right\|_2 \cdot \textbf{M}^{i}),
  \label{eq:eq_def_div}
  \vspace{-3mm}
\end{equation}
where $\mathbf{R}^{correct}$ represents an objectively existing correct feature, and $\mathbf{M}^{i}$ is an instance mask generated according to the box $\mathbf{b}^i$, with the values inside the box region are set to 1, and the rest of the positions are set to 0. That is to say, each shading instance should have the correct textual semantic in its corresponding area.

\noindent \textbf{Two benefits of division in the Cross-Attention space.} i.e., 1) Conquer more efficiently: For N-instance generation, MIGC conquers N subtasks solely on Cross-Attention layers instead of the entire Unet network, which will be more efficient; 2) Combine more harmoniously: Combining subtasks in the middle layer enhances the overall cohesiveness of the generated image compared to combining at the final output of the network. 

\definecolor{mygray2}{gray}{.8}
\begin{table*}[!htb]
\setlength\tabcolsep{4pt}
\setlength{\abovecaptionskip}{0.cm}
    \setlength{\belowcaptionskip}{0.cm}
  \centering
  % \begin{tabular}{@{}lc @{}c @{}c @{}c @{}c @{}c @{}c @{}c @{}c }
  \begin{tabular}{c| c  c c c c  >{\columncolor{mygray2}}c | c c c c c >{\columncolor{mygray2}}c | c  }
    \toprule
    \multicolumn{1}{c|}{\textbf{Method}} & \multicolumn{6}{c|}{\textbf{Instance Success Rate(\%)}$\uparrow$}  & \multicolumn{6}{c|}{\textbf{mIoU$\uparrow$}} & \multirow{2}{*}{\textbf{Time(s)$\downarrow$}} \\

    \cmidrule(lr){1-1} \cmidrule(lr){2-7} \cmidrule(lr){8-13} 
    
    Level & ${L_2}$ & ${L_3}$ & ${L_4}$ & ${L_5}$ & ${L_6}$ & Avg  & ${L_2}$ & ${L_3}$ & ${L_4}$ & ${L_5}$ & ${L_6}$ & Avg  \\
    \midrule
    Stable Diffusion & 6.87 & 5.01 & 3.45 & 3.27 & 2.21 & 3.61 & 18.92 & 17.44 & 15.85 & 15.17 & 14.42 & 15.80 & \textbf{9.18}  \\
    TFLCG & 20.47 & 12.71 & 8.36 & 6.72 & 4.36 & 8.62 & 29.34 & 25.06 & 20.82 & 18.81 & 17.86 & 20.92 & 19.92   \\
    BOX-Diffusion &  24.61 & 19.22 & 14.20 & 11.92 & 9.31 & 13.96 & 32.64 & 29.88 & 25.39 & 23.81 & 21.19 & 25.14 & 44.17   \\
    Multi Diffusion & 24.88 & 22.14 & 19.88 & 18.97 & 18.60 & 20.12 & 29.41 & 28.06 & 25.59 & 24.83 & 24.71 & 25.89 & 25.15  \\
    GLIGEN & 42.30 & 35.55 & 32.66 & 28.18 & 30.84 & 32.39 & 37.58 & 32.34 & 29.95 & 26.60 & 27.70 & 32.25 & 22.00   \\
    \midrule
    Ours & \textbf{67.70} & \textbf{59.61} & \textbf{58.09} & \textbf{56.16} & \textbf{56.88} & \textbf{58.43} & \textbf{59.39} & \textbf{52.73} & \textbf{51.45} & \textbf{49.52} & \textbf{49.89} & \textbf{51.48} & 15.61  \\
    \bottomrule
  \end{tabular}
  \caption{ Quantitative results in our proposed COCO-MIG benchmark. According to the count of generated instances, COCO-MIG is divided into five levels: $L_{2}$, $L_{3}$, $L_{4}$, $L_{5}$, and $L_{6}$. $L_{i}$ means that the count of instances needed to generate in the image is i. }
  \label{tab:COCO-MIG}
\end{table*}

\begin{figure*}[tb]
    % \vspace{-2mm}
    \setlength{\abovecaptionskip}{0.cm}
    \setlength{\belowcaptionskip}{0.cm}
    \begin{center}
        \includegraphics[width=1.0\textwidth]{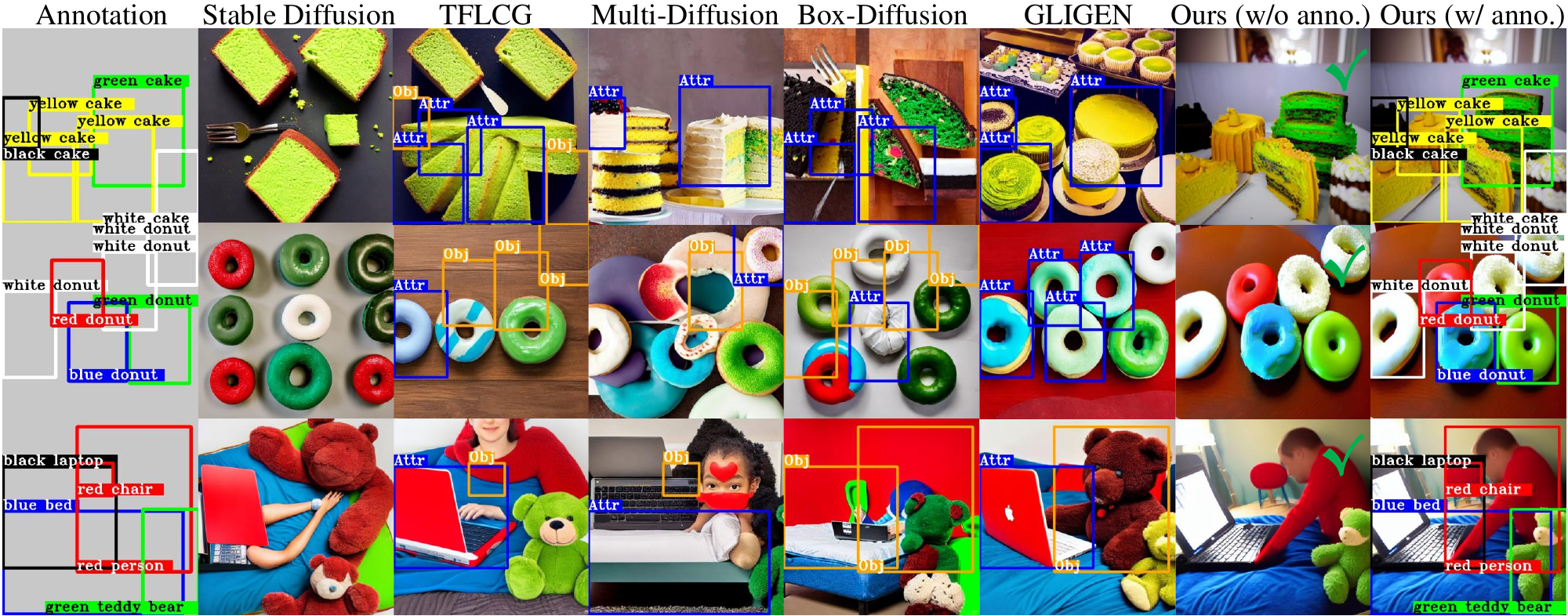}
    % \vspace{-1mm}
    \end{center}
    
   \caption{Qualitative comparison of our MIGC and other baselines on COCO-MIG. We use a yellow bounding box labeled ``Obj" to indicate a position-wrongly generated instance and a blue bounding box labeled ``Attr" to indicate an attribute-wrongly generated instance. Experimental results show that MIGC can achieve better attribute (i.e., color) control while precisely controlling the positions of instances.} 
  \vspace{-1em}
    \label{fig:result_color}
  \end{figure*}  

\subsection{Conquer Instance Shading}

\noindent \textbf{Shading stage 1: shading results of Cross-Attention.} The pre-trained Cross-Attention will notice regions with high attention weight and perform shading according to the textual semantics~\cite{diffumask,diffedit}. As shown in Fig.~\ref{fig:fig_2}, MIGC uses the masked Cross-Attention output as the first shading results:
\vspace{-0.2em}
\begin{equation}
\vspace{-0.2em}
  \mathbf{R}^{i}_{f} = \text{Softmax}(\frac{\mathbf{Q} \mathbf{K^{i}}^T}{\sqrt{d}}) \mathbf{V^{i}} \cdot \textbf{M}^{i},
  \label{eq:eq_R_pre}
\end{equation}
where $\mathbf{K}^{i}$ and $\mathbf{V}^{i}$ are obtained from text embedding of $\mathbf{d}^i$, and $\mathbf{Q}$ is obtained from the image feature map.

\noindent \textbf{Two issues of Cross-Attention shading results.} 1) \textit{Instance Merge.} According to Eq.~\eqref{eq:eq_R_pre}, for two instances with the same description, they will get the same $\mathbf{K}$ and $\mathbf{V}$ in the Cross-Attention layer. If their boxes are close or even overlap, the network will easily merge the two instances; 2) \textit{Instance Missing.} The initial edit method~\cite{initialimageedit} shows that the initial noise of SD largely determines the layout of the generated image, i.e., specific regions prefer to generate specific instances or nothing. If the initial noise does not tend to generate an instance according to description $\mathbf{d}^{i}$ in box $\mathbf{b}^i$, the $\mathbf{R}_{f}^{i}$ will be weak, leading to the instance missing.

\noindent
\textbf{Grounded phrase token for solving instance merge.} 
To identify instances with the same description but different boxes, MIGC extends the text tokens of each instance to a combination of text and position tokens. As shown in Fig.~\ref{fig:fig_overview}(a),  
MIGC first projects the bounding box information to the Fourier embedding, then uses a MLP layer to get position tokens. MIGC concatenates the text tokens with position tokens to obtain the grounded phrase tokens:
\vspace{-0.2em}
\begin{equation}
\vspace{-0.2em}
  \mathbf{G}^{i} = [\text{CLIP}(\mathbf{d}^{i}),\text{MLP}(\text{Fourier}(\mathbf{b}^{i}))] ,
  \label{eq:eq_ground_token}
\end{equation}
where $[\cdot]$ represents the concatenation. 

\noindent
\textbf{Shading stage 2: Enhancement Attention for solving instance missing.} Illustrated in Fig.~\ref{fig:fig_2}, MIGC uses a trainable Enhancement-Attention (EA) Layer to enhance the shading result. Specifically, as shown in Fig.~\ref{fig:fig_overview}(a), after obtaining the grounded phrase token, EA uses a new trainable Cross-Attention layer to obtain an enhanced shading result and adds it to the first shading result $\mathbf{R}^{i}_{f}$:

\vspace{-0.5em}
\begin{equation}
  \mathbf{R}^{i}_{s} = \mathbf{R}^{i}_{f} + \text{softmax}(\frac{\mathbf{Q}_{ea} {\mathbf{K}^{i}_{ea}}^T}{\sqrt{d}}) \mathbf{V}^{i}_{ea} \cdot \textbf{M}^{i},
  \label{eq:eq_R_aug1}
\end{equation}
where $\mathbf{K}^{i}_{ea}$ and $\mathbf{V}^{i}_{ea}$ are obtained from the grounded phrase token $\mathbf{G}^i$, and $\mathbf{Q}_{ea}$ is obtained from the image feature map.
During the training period, since $\mathbf{M}^{i}$ ensures precise spatial positioning, the instance shading result output by EA exclusively impacts the correct region, so the EA can easily learn: no matter what the image feature is, the EA should perform enhanced shading to satisfy the textual semantic of $\mathbf{d}^{i}$ and solve the issue of instance missing.

\textit{Finally, MIGC treats the second shading result $\mathbf{R}^{i}_{s}$ as the solution of the $subtask_{i}$.}

\noindent

\subsection{Combine Shading Results}

\noindent
\textbf{Global prompt residual as shading background.} 
 Obtaining N-instance shading results as shading foreground, the next step of MIGC is to get the shading background. Illustrated in Fig.~\ref{fig:fig_2}(c), MIGC utilizes global prompt $\mathcal{P}$ to obtain the shading background result $\mathbf{R}^{bg}$ in a manner similar to Eq.\eqref{eq:eq_R_pre}, with the background mask $\mathbf{M}^{bg}$, in which positions containing the instance are assigned a value of 0, while all other positions are marked as 1.
\vspace{-0.2em}

\noindent
\textbf{Layout Attention residuals as shading template.} 
A certain gap exists between shading instances $\{\mathbf{R}_{s}^{1},\ldots,\mathbf{R}_{s}^{N}\}$ and the shading background $\mathbf{R}^{bg}$, as their shading process is relatively independent. To bridge these shading results and minimize the gap, MIGC needs to learn a shading template according to the image feature maps' information.
As shown in Fig.~\ref{fig:fig_2}(c), a Layout Attention layer is used in MIGC to achieve the above goal. Illustrated in Fig.~\ref{fig:fig_overview}(b), Layout Attention performs similarly to the Self-Attention~\cite{selfattn,palette} while instance masks $\mathbb{M}_{inst}=\{\mathbf{M}^{bg},\mathbf{M}^{1}, \ldots, \mathbf{M}^{N}\}$ are used to construct attention masks:
\vspace{-0.2em}
\begin{equation}
\vspace{-0.8em}
\mathbf{A}_{(a,b),(c,d)} = \begin{cases}
1,  \text{if}\, \exists \ \mathbf{m} \in \mathbb{M}_{inst},\mathbf{m}_{a,b}= \mathbf{m}_{c,d}=1 \\
-inf, \ \ \ \ \ \  \text{otherwise}
\end{cases}
  \label{eq:eq_la_mask}
\end{equation}

\begin{equation}
  \mathbf{R}_{LA} = \text{Softmax}(\frac{\mathbf{Q}_{LA} {\mathbf{K}_{LA}}^T}{\sqrt{d}} \odot \mathbf{A})\mathbf{V}_{LA},
  \label{eq:eq_RLA}
\end{equation}
where $\odot$ represents the Hadamard product, and $\mathbf{A} \in \mathbb{R}^{((H,W),(H,W))}$ represents attention masks, in which $\mathbf{A}_{(a,b)(c,d)}$ determines whether pixel (a, b) should attend to pixel (c, d). The constructed attention mask $\textbf{A}$ ensures one pixel can only attend to other pixels in the same instance region, which avoids attribute leakage between instances.

\noindent
\textbf{Shading Aggregation Controller for the final fusion.} 
To summarize, in all the above operations, MIGC can get $\mathbb{R}_s=\{\mathbf{R}_{s}^{1},\ldots,\mathbf{R}^{N}_{s}, \mathbf{R}^{bg},\mathbf{R}_{LA}\}\in \mathbb{R}^{(N+2,C,H,W)}$ and $\mathbb{M}=\{\mathbf{M}^{1},\ldots,\mathbf{M}^{N}, \mathbf{M}^{bg},\mathbf{M}_{LA}\}\in \mathbb{R}^{(N+2,1,H,W)}$, where $\mathbf{M}_{LA}$ is the all-1 guidance mask corresponding to $\mathbf{R}_{LA}$. In order to dynamically aggregate shading results at different timesteps of the generation process, we propose the Shading Aggregation Controller (SAC). As shown in Fig.\ref{fig:fig_overview}(c), 
SAC sequentially performs instance intra-attention and inter-attention, and aggregation weights summing to 1 are assigned to shading results on each spacial pixel through the softmax function, resulting in the final shading.
\begin{equation}
\mathbf{R}_{final}=SAC(\mathbb{R}_s, \mathbb{M}), \mathbf{R}_{final} \in \mathbb{R}^{H,W,C}
  \label{eq:eq_final_res}
\end{equation}
For more details, please refer to supplementary materials.
%-------------------------------------------------------------------------

%-------------------------------------------------------------------------

\definecolor{mygray2}{gray}{.8}
\begin{table*}[h!tbp]
\setlength\tabcolsep{6pt}
\setlength{\abovecaptionskip}{0.cm}
    \setlength{\belowcaptionskip}{0.cm}
  \centering
  % \begin{tabular}{@{}lc @{}c @{}c @{}c @{}c @{}c @{}c @{}c @{}c }
  \begin{tabular}{c| c c c c c | c c | c }
    \toprule
    \multirow{2}{*}{\textbf{Method}} & \multicolumn{5}{c|}{\textbf{Spatial Accuracy(\%)}}  & \multicolumn{2}{c|}{\textbf{Image Text Consistency}}  & \multicolumn{1}{c}{\textbf{Image Quality}}\\

    \cmidrule(lr){2-6} \cmidrule(lr){7-8} \cmidrule(lr){9-9} 
    
    & R$\uparrow$ & mIoU$\uparrow$ & AP$\uparrow$ & AP50$\uparrow$ & AP75$\uparrow$ & CLIP$\uparrow$ & Local CLIP$\uparrow$ & FID-6K$\downarrow$  \\
    \midrule
    \rowcolor{mygray2}Real Image & 83.75 & 85.49 & 65.97 & 79.11 & 71.22 & 24.22 & 19.74 & -  \\
    \midrule
    Stable Diffusion & 5.95 & 21.60 & 0.8 & 2.71 & 0.42 & \textbf{25.69} & 17.34 & \textbf{23.56}  \\
    TFLCG & 13.54 & 28.01 & 1.75 & 6.77 & 0.56 & 25.07 & 17.97 & 24.65  \\
    BOX-Diffusion & 17.84 & 33.38 & 3.29 & 12.27 & 1.08 & 23.79 & 18.70 & 25.15  \\
    Multi Diffusion & 23.86 & 38,82 & 6.72 & 18.65 & 3.63 & 22.10 & 19.13 & 33.20    \\
    Layout Diffusion & 50.53 & 57.49 & 23.45 & 48.10 & 20.70 & 18.28 & 19.08 & 25.94  \\
    GLIGEN & 70.52 & 71.61 & 40.68 & 68.26 & 42.85 & 24.61 & 19.69 & 26.80   \\
    % ReCo & 73.22 & \textbf{73.77} & \textbf{53.28} & \textbf{82.36} & \textbf{60.68} & 23.00 & \textbf{20.74} & 66.92 \\
    \midrule
    Ours & \textbf{80.29} & \textbf{77.38} & \textbf{54.69} & \textbf{84.17} & \textbf{61.71} & 24.66 & \textbf{20.25} & 24.52   \\
    \bottomrule
  \end{tabular}
  \caption{Quantitative results on the COCO-Position. ``R'' means the success rate, which checks whether all instances in one image are position-correctly generated.}
  \vspace{-1em}
  \label{tab:COCO-Position}
\end{table*}

\begin{figure*}[tb]
    % \vspace{-2mm}
    \setlength{\abovecaptionskip}{0.cm}
    \setlength{\belowcaptionskip}{0.cm}
    \begin{center}
        \includegraphics[width=1.0\textwidth]{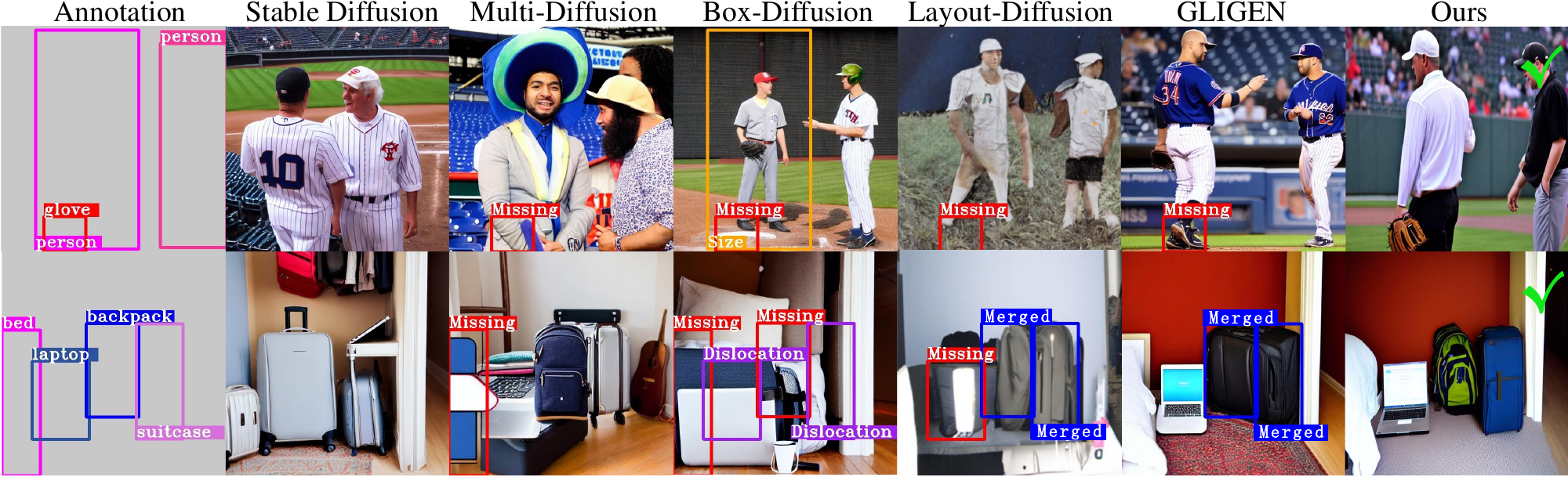}
    % \vspace{-1mm}
    \end{center}
    
   \caption{Qualitative comparison of MIGC and other baselines on COCO-Position. 
   % In the image results generated by the baseline method, we use four types of bounding boxes to identify different error situations. The red ``Missing" box indicates \textbf{a missing object}, the yellow ``Size" box denotes a \textbf{significant size mismatch} between the generated object and the bounding box, the blue ``Merged" label indicates two bounding boxes where \textbf{objects have merged into one (instance merging)}, and the ``Dislocation" label signifies a \textbf{displacement between the generated object's current position and its actual position}. 
   Experimental results show that our method can reduce the problem of instance missing, improve positional control, and alleviate the phenomenon of instance merging.
   }
   \vspace{-1em}
    \label{fig:result_position}
  \end{figure*}

\subsection{Summary}

\noindent
\textbf{Training Loss.} We use the original denoising loss \cite{stablediffusion,ddpm}:

\vspace{-0.5em}
\begin{equation}
  \min_{\theta'} \mathcal{L}_{\text{LDM}} = \mathbb{E}_{z,\epsilon \sim \mathcal{N}(0,I),t}[||\epsilon - f_{\theta,\theta'}(\mathbf{z}_t,t,\mathcal{P},\mathbb{B},\mathbb{D})||_2^2],
  \label{eq:eq_14}
\end{equation}
where $\theta$ represents the frozen parameters of the pre-trained stable diffusion, and $\theta'$ means the parameter of our MIGC.

Besides, to constrain generated instances within their regions and prevent the generation of additional objects in the background, we design an inhibition loss to avoid high attention weight in the background region:

\vspace{-0.8em}
\begin{equation}
\vspace{-0.2em}
  \min_{\theta'} \mathcal{L}_{\text{ihbt}} = \sum_{i=1}^{i=N} \left |\mathbf{A}_c^i - \text{DNR}(\mathbf{A}_c^i) \right | \odot \mathbf{M}^{bg} ,
  \label{eq:eq_15}
\end{equation}
where $\mathbf{A}_c^i$ denotes the attention maps for the ith instance in the frozen $16\times{16}$ Cross-Attention layer of the Unet decoder~\cite{initialimageedit}, and DNR($\cdot$) means the denoising (e.g., we use the average operation) of the background region. The final training loss is designed as follows:

\vspace{-0.2em}
\begin{equation}
\vspace{-0.2em}
  \min_{\theta'} \mathcal{L} = \mathcal{L}_{\text{LDM}} + \lambda \mathcal{L}_{\text{ihbt}},
  \label{eq:eq_14}
\end{equation}
we set the loss weight $\lambda$ as 0.1.

\noindent
\textbf{Implementation Details.} We only deploy MIGC on the mid-layers (i.e., $8\times{8}$) and the lowest-resolution decoder layers (i.e., $16\times{16}$) of UNet, which greatly determine the generated image's layout and semantic information~\cite{dragondiffusion,tflcg}. In other Cross-Attention layers, we use the global prompt for global shading. We use COCO 2014 \cite{coco} to train MIGC. To get the instance descriptions and their bounding boxes, we use stanza \cite{stanza} to split the global prompt and detect the instances with the Grounding-DINO\cite{gdino} model. We train our MIGC based on the pre-trained stable diffusion v1.4. We use AdamW \cite{adam} optimizer with a constant learning rate of $1e^{-4}$, and train the model for 300 epochs with batch size 320, which requires 15 hours on 40 V100 GPUs with 16GB VRAM each. 
For inference, we use EulerDiscreteScheduler~\cite{elud} with 50 sample steps and use our MIGC in the first 25 steps. We select the CFG scale\cite{Ho2022ClassifierFreeDG} as 7.5. For more details, please refer to supplementary materials.

\section{Experiments}
\label{sec:formatting}
\newcommand{\imporve}{\textcolor{green}}
\newcommand{\down}{\textcolor{red}}

\subsection{Benchmarks}
\label{benchmark}

We evaluate models' performance on three benchmarks: COCO-MIG, COCO-Position~\cite{coco}, and DrawBench~\cite{imagen}. We use 8 seeds to generate images for each prompt.

\noindent
\textbf{In COCO-MIG}, we pay attention to position, color, and quantity. To construct it, we randomly sampled 800 COCO images and assigned a color to each instance while keeping the original layout. Furthermore, We reconstruct the global prompts in the format of ' a $<$attr1$>$ $<$obj1$>$ and a $<$attr2$>$ $<$obj2$>$ and a ... ', and we divide this benchmark into five levels based on the number of instances in the generated image. Each method will generate 6400 images.

\noindent
\textbf{In COCO-Position}, we sampled 800 images, using the captions as the global prompts, labels as instance descriptions, and bounding boxes as layouts to generate 6400 images.

\noindent
\textbf{Drawbench} is a challenging T2I benchmark. We use GPT4 \cite{gpt4,layoutgpt} to extract all instance descriptions and generate the layouts for each prompt. We use a total of 64 prompts, of which 25 are related to color, 19 are related to counting, and 20 are related to position, ultimately generating 512 images.

%-------------------------------------------------------------------------
\subsection{Evaluation Metrics}
\label{metric}

\noindent
\textbf{Position Evaluation.} We use Grounding-DINO~\cite{gdino} to detect each instance and calculate the maximum IoU between the detection boxes and the Ground Truth box. If the above IoU is higher than the threshold $t$=0.5, we mark it as \textit{Position Correctly Generated}.

\noindent
\textbf{Attribute Evaluation.} For a \textit{Position Correctly Generated} instance, we use the Grounded-SAM model~\cite{gdino,sam} to segment it and calculate the percentage of the target color in the HSV color space.  If the above percentage exceeds the threshold $S$=0.2, we denote it as \textit{Fully Correctly Generated}.

\noindent
\textbf{Metrics on COCO-MIG.} We primarily measure the \textit{Instance Success Rate} and mIoU. The \textit{Instance Success Rate} calculates the probability that \textbf{each instance} is \textit{Fully Correctly Generated}, and mIoU calculates the mean of the maximum IoU for all instances. Note that if the color attribute is incorrect, we set \textbf{the IoU value as 0}.

\noindent
\textbf{Metrics on COCO-Position.} We use \textit{Success Rate}, mIoU and Grounding-DINO AP score to measure the Spatial Accuracy. The \textit{Success Rate} represents whether \textbf{all instances} in one image are \textit{Position Correctly Generated}. Besides, we use the Fr\'echet Inception Distance (FID) \cite{FID} to evaluate Image Quality. To measure Image-Text Consistency, we use CLIP score and Local CLIP score\cite{spatext}.

\noindent
\textbf{Metrics on DrawBench.}
We evaluate the \textit{Success Rate} for images related to the position and count by checking whether all instances in each image are \textit{Position Correctly Generated}. For color-related images, we check whether all instances are \textit{Fully Correctly Generated}. In addition to automated evaluations, a manual evaluation is conducted.
%-------------------------------------------------------------------------
\definecolor{mygray2}{gray}{.8}
\begin{table}[tb]
\setlength{\abovecaptionskip}{0.cm}
\setlength{\belowcaptionskip}{0.cm}
\setlength\tabcolsep{1.5pt}
  \centering
  % \begin{tabular}{@{}lc @{}c @{}c @{}c @{}c @{}c @{}c @{}c @{}c }
  \begin{tabular}{c| c c | c c | c c  }
    \toprule
    \multirow{2}{*}{\textbf{Method}} & \multicolumn{2}{c|}{\textbf{Spatial(\%) $\uparrow$}}  & \multicolumn{2}{c|}{\textbf{Attribute(\%) $\uparrow$}}  & \multicolumn{2}{c}{\textbf{Count(\%) $\uparrow$}} \\

    \cmidrule(lr){2-3} \cmidrule(lr){4-5} \cmidrule(lr){6-7} 
    
    & R  & Human & R & Human & R & Human  \\
    \midrule
    SD1.4 & - & 13.30 & - & 57.52 & - & 23.70 \\
    AAE & - & 23.13 & - & 51.50 & - & 30.92 \\
    Struc-D & - & 13.12 & - & 56.5 & - & 30.26 \\
    \midrule
    Box-D & 11.88 & 50.00 & 28.50 & 57.50 & 9.21 & 39.47   \\
    TFLCG & 9.38 & 53.13 & 35.00 & 60.00 & 15.79 & 31.58    \\
    Multi-D & 10.63 & 55.63 & 18.5 & 65.50 & 17.76 & 36.18    \\
    GLIGEN & 61.25 & 78.80 & 51.00 & 48.20 & 44.08 & 55.90  \\
    % ReCo & 68.75 & 80.00 & 79.00 & 89.00 & 32.24 & 46.67  \\
    \midrule
    Ours & \textbf{69.38} & \textbf{93.13} & \textbf{79.00} & \textbf{97.50} & \textbf{67.76} & \textbf{67.50}   \\
    \bottomrule
  \end{tabular}
  \caption{Evaluation on the drawbench.}
  \vspace{-1em}
  \label{tab:DrawBench}
\end{table}

% \vspace{-1.7mm}
\subsection{Baselines}
\label{baseline}

We compare our method with some SOTA layout-to-image methods:  Multi-Diffusion\cite{multidiffusion}, Layout Diffusion\cite{layoutdiff}, GLIGEN\cite{gligen}, TFLCG\cite{tflcg}, and Box-Diffusion\cite{boxdiff}. Since Layout Diffusion cannot control color, we only run it on COCO-Position. In Drawbench, we also compare our method with some SOTA T2I methods: stable diffusion v1.4\cite{stablediffusion}, AAE\cite{aae}, Structure Diffusion\cite{structurediff}. All methods are executed using the official code and default configuration.

 % \vspace{-2mm}

\begin{figure}[tb]
\setlength{\abovecaptionskip}{0.cm}
\setlength{\belowcaptionskip}{0.cm}
    \begin{center}
        \includegraphics[width=0.45\textwidth]{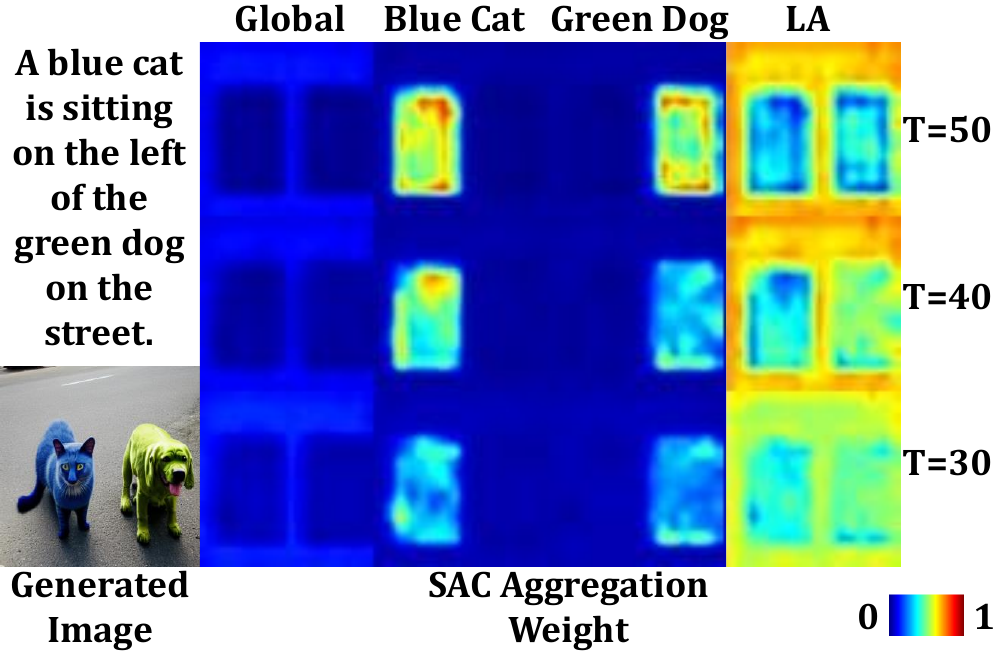}
        
    \end{center}
   % \vspace{-4mm}
    \caption{SAC aggregation weight at T=50, 40, and 30. T=50 means the first step, as we generated each image with 50 steps.} 
    \vspace{-1mm}
    \label{fig:fusion_weight}
  \end{figure}

\subsection{Quantitative Results}
\label{quantitative}

\noindent
\textbf{COCO-MIG.} Tab.\ref{tab:COCO-MIG} shows results in COCO-MIG. MIGC improves the \textit{Instance Success Rate} from 32.39\% to 58.43\% and mIoU from 32.25 to 51.48. Improvements are consistently observed across all count-division levels, underscoring the robust control capabilities of MIGC on position, quantity, and attributes. Furthermore, MIGC runs at almost the same speed as the original stable diffusion, thanks to MIGC dividing MIG in the Cross-Attention Space, accelerating the conquering and combing of subtasks.

% \vspace{-2mm}

% 超过原图 该如何说明
\noindent
\textbf{COCO-Position.} Tab.\ref{tab:COCO-Position} shows quantitative results in COCO-Position, indicating that MIGC brings significant improvement in Spatial Accuracy: increased the \textit{Success Rate} from 70.52\% to 80.29\%,  mIoU from 71.61 to 77.38, and AP score from 40.68/68.26/42.85 to 54.69/84.17/61.71.  MIGC also achieves similar FID scores compared to the stable diffusion, highlighting that MIGC can enhance position control capabilities without destroying image quality.

\noindent
\textbf{DrawBench.} Tab.\ref{tab:DrawBench} shows the results in drawbench. MIGC achieves the best performance in both mechanical metrics and human evaluation. Human evaluation doesn't rely on IoU to determine position correctness.

\begin{table}[htb]
  \centering
  \setlength\tabcolsep{1.2pt}
  \setlength{\abovecaptionskip}{0.cm}
    \setlength{\belowcaptionskip}{0.cm}
  \begin{tabular}{c c c | c c c c c | c  }
    \toprule
   \textbf{SAC} & \textbf{EA} & \textbf{LA} & \textbf{R}(\%)$\uparrow$  & \textbf{mIoU}$\uparrow$ & \textbf{AP}$\uparrow$ &  \textbf{AP50}$\uparrow$ & \textbf{AP75}$\uparrow$ & \textbf{numb.}  \\
    \midrule
              &           &            & 7.66 & 22.71  & 0.91 & 3.18 & 0.35  & \ding{172}  \\
    \ding{52} &           &            & 12.10 & 29.55 & 1.89 & 7.64 & 0.49  & \ding{173}  \\ 
    \ding{52} &           & \ding{52}  & 34.70 & 44.08 & 11.02 & 28.64 & 6.83 & \ding{174}  \\
    \ding{52} & \ding{52} &            & 80.16 & 76.63 & 53.03 & 84.05 & 58.67 & \ding{175}  \\
              & \ding{52} & \ding{52}  & 78.12 & 75.47 & 52.05 & 83.48 & 57.16 & \ding{176}  \\
    \midrule
    \ding{52} & \ding{52} & \ding{52}  & \textbf{80.29} & \textbf{77.38} & \textbf{54.69}& \textbf{84.17} & \textbf{61.71} & \ding{177}  \\
    \bottomrule
  \end{tabular}
  \caption{Ablation on COCO-Position of \textbf{Shading Aggregation Controller(SAC), Enhancement Attention (EA), Layout Attention (LA)}. }
  \vspace{-1em}
  \label{tab:ablation}
\end{table}

\begin{table}[htb]
% \vspace{-2em}
  \centering
  \setlength\tabcolsep{1.5pt}
  \setlength{\abovecaptionskip}{0.cm}
    \setlength{\belowcaptionskip}{0.cm}
  \begin{tabular}{c | c c c c c c  }
    \toprule
   \textbf{Config} & \textbf{R}(\%)$\uparrow$  & \textbf{mIoU}$\uparrow$ & \textbf{AP}$\uparrow$ & \textbf{AP50}$\uparrow$ & \textbf{AP75}$\uparrow$ & \textbf{FID}$\downarrow$ \\
    \midrule
    w/o loss & 80.20 & 77.03  & 52.46 & 82.65 & 58.05 & 24.73   \\
    w/ loss 1.0 & \textbf{80.61} & \textbf{77.79}  & \textbf{55.62} & \textbf{84.48} & \textbf{62.85} & 26.94    \\
    w/ loss 0.1 & 80.29 & 77.38 & 54.69 & 84.17 & 61.71 & \textbf{24.52}  \\ 
    \bottomrule
  \end{tabular}
  \caption{Ablation on COCO-Position of \textbf{Inhibition loss}. We conducted an ablation study on three configurations: w/o loss, loss weight 1.0, and loss weight 0.1. }
  \vspace{-2mm}
  \label{tab:ablation_loss}
\end{table}

\subsection{Qualitative Results}
\label{qualitative}
Fig.\ref{fig:result_color} shows qualitative results in COCO-MIG. MIGC demonstrates effective position-and-attributes control over all instances, even in complex scenarios. Fig.\ref{fig:result_position} shows qualitative results in COCO-Position. MIGC achieves more precise control, ensuring all instances are generated strictly within their designated boxes without instances missing or merging. The qualitative results for DrawBench will be presented in the supplementary materials.

\subsection{Analysis of Shading Aggregation Controller}

We generate each image with 50 steps while using MIGC in the first 25 steps. Fig.\ref{fig:fusion_weight} shows SAC aggregation weights at T=50, 40, and 30 (i.e., T=50 means the first step). In the early time steps, the SAC assigns more weight to the EA layer's shading instances in the foreground while giving more weight to the LA layer's shading template in the background. In the later time steps, the SAC gradually increases the attention to the global context in the background.

\subsection{Ablation Study}
\label{ablation}
The ablation focuses on four components: (1) Enhancement Attention Layer. (2) Layout Attention Layer. (3) Shading Aggregation Controller. (4) The inhibition loss. Experiments are performed on COCO-Position and COCO-MIG.

\begin{figure}[tb]
    \setlength{\abovecaptionskip}{-0.cm}
    \setlength{\belowcaptionskip}{-0.cm}
    \begin{center}
        \includegraphics[width=0.45\textwidth]{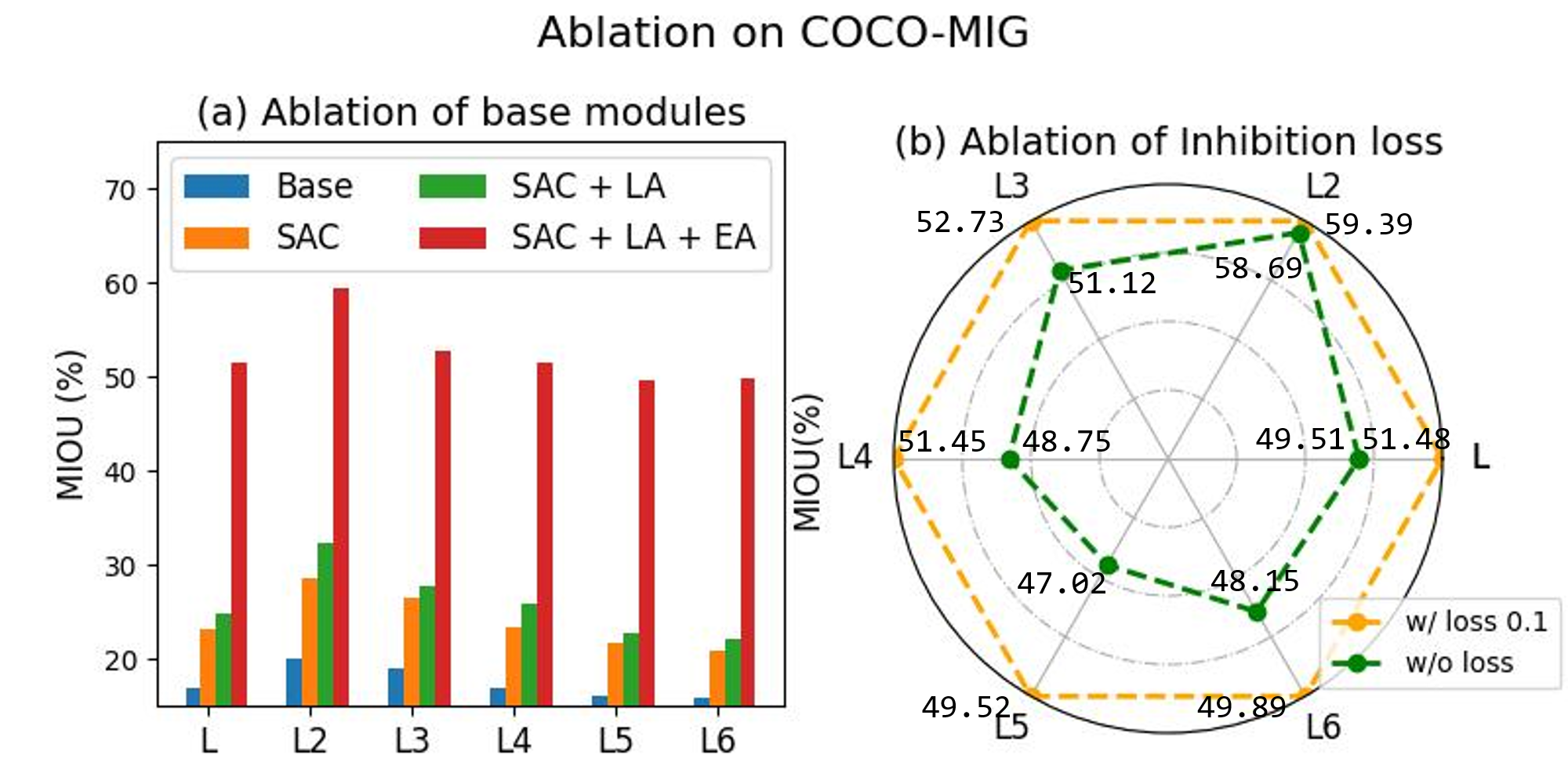}
        
    \end{center}
   % \vspace{-4mm}
    \caption{The results of ablation studies on COCO-MIG. (a) shows the ablation results of three components, and (b) shows the ablation results of inhibition loss.  } 

    \label{fig:ablation_MIG}
  \end{figure}

\noindent
\textbf{Shading Aggregation Controller.}
From Tab.\ref{tab:ablation}, we find that using SAC improves the performance metrics(compare \ding{176} with \ding{177} and \ding{172} with \ding{173}), which is also reflected in the ablation experiments on COCO-MIG in Fig.\ref{fig:ablation_MIG}(a).

\noindent
\textbf{Enhancement Attention Layer.}
In Tab.\ref{tab:ablation}, the EA Layer significantly improves the \textit{Success Rate} from 12.10\% to 80.16\%, mIoU from 29.55 to 76.63, and AP from 1.89 / 7.64 / 0.49 to 53.03 / 84.05 / 58.67 (Compare \ding{173} with \ding{175}). We also observe significant improvement in Fig.\ref{fig:ablation_MIG}(a).

\begin{figure}[tb]
\vspace{-2mm}
    \setlength{\abovecaptionskip}{-0.cm}
    \setlength{\belowcaptionskip}{-0.cm}
    \begin{center}
        \includegraphics[width=0.47\textwidth]{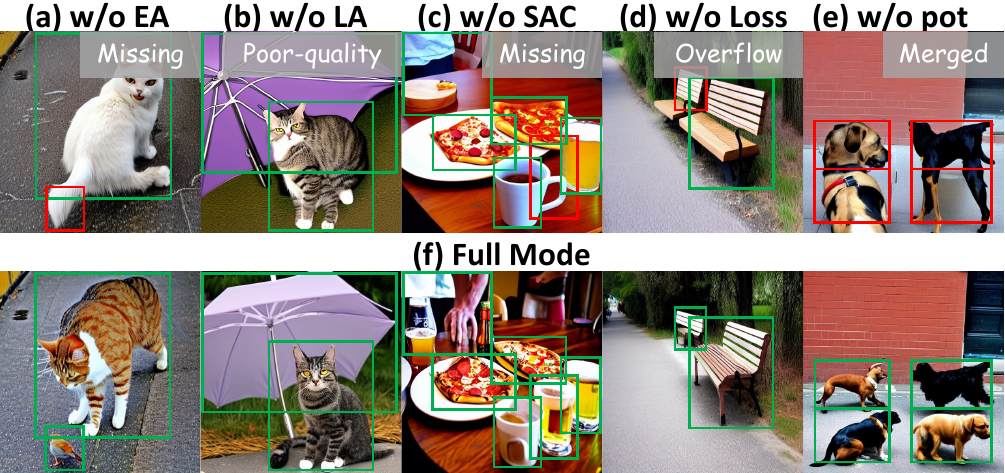}
        
    \end{center}
   % \vspace{-4mm}
    \caption{ The qualitative results of ablation studies. Pot means the Position Token used in EA Layer. We mark incorrectly generated instances with \textcolor{red}{red} boxes and correctly with \textcolor{green}{green} boxes. } 
\vspace{-1em}
    \label{fig:ablation_qualitative}
  \end{figure}

\noindent
\textbf{Layout Attention Layer.}
The results of \ding{175} and \ding{177} in Tab.\ref{tab:ablation} show that LA Layer can improve the AP. We find that SAC+LA, compared to SAC alone, has improved mIoU to some extent in Fig.\ref{fig:ablation_MIG}(a).

\noindent
\textbf{Inhibition Loss.}
We also conducted the ablation study on the Inhibition loss, with 0.1 and 1.0 loss weight. We show the results in Tab.\ref{tab:ablation_loss} and Fig.\ref{fig:ablation_MIG}(b). Tab.\ref{tab:ablation_loss} indicates that inhibition loss can significantly improve the AP metric in COCO-Position. We find that setting the loss function weight to 1.0 can further improve the AP metric, but it comes at the cost of a slight decrease in image quality (i.e., FID). So we finally choose loss weight as 0.1. Fig.\ref{fig:ablation_MIG} (b) shows the comparison between \textit{w/ loss 0.1} and \textit{w/o loss} on COCO-MIG, and we observe that the inhibition loss can improve the mIoU, especially when generating images with large instance quantity.

\noindent
\textbf{Qualitative Results.}
We show qualitative results in Fig.\ref{fig:ablation_qualitative}. The first column indicates that the EA Layer can effectively alleviate instance missing. The second column illustrates that the LA Layer can significantly improve generated image quality. The third column suggests that the SAC also aids in better aggregation of shading instances. The fourth column demonstrates that inhibition loss enhances the model's control capabilities. The fifth column demonstrates that position tokens effectively alleviate instance merging.

\section{Conclusion}
\label{sec:formatting}

In this work, we define a practical and challenging MIG task and propose a MIGC approach to improve the stable diffusion's MIG ability. We divide the complex MIG task into simpler Single-Instance shading subtasks, conquer each instance shading with an Enhancement Attention layer, and combine the final shading result through a Layout Attention layer and Shading Aggregation Controller. Comprehensive experiments are conducted on our proposed COCO-MIG and popular COCO-Position and Drawbench benchmarks. Experiment results verify the efficiency and effectiveness of our MIGC. In the future, we will further explore the control of interactive relationships between instances.

{
    \small
    \bibliographystyle{ieeenat_fullname}
    \bibliography{main}
}

\clearpage
\setcounter{page}{1}
\setcounter{table}{0}
\setcounter{figure}{0}
\setcounter{section}{0}
\setcounter{equation}{0}

\renewcommand{\thesection}{\Alph{section}}

\maketitlesupplementary

\begin{figure}[tb]
    % \vspace{-2mm}
    \setlength{\abovecaptionskip}{0.cm}
    \setlength{\belowcaptionskip}{0.cm}
    \begin{center}
        \includegraphics[width=0.45\textwidth]{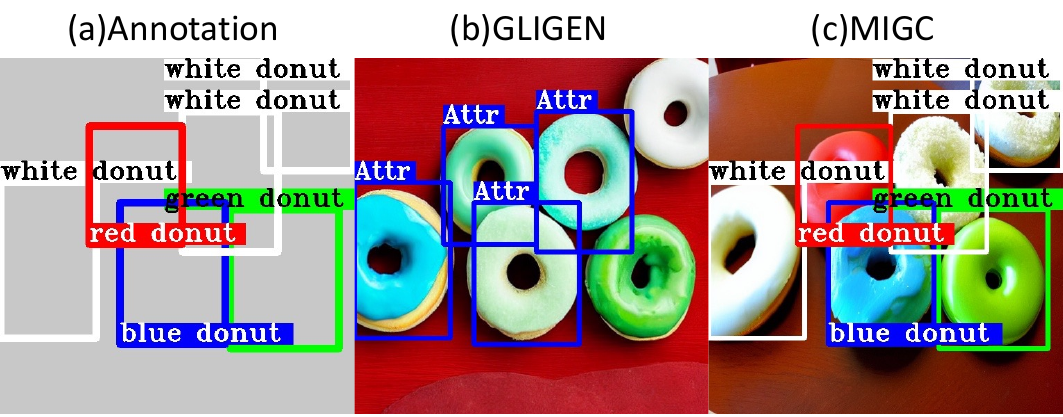}
    % \vspace{-1mm}
    \end{center}
   \caption{An example from the COCO-MIG benchmark. (a) In this example, COCO-MIG requires generation models to generate ``donuts" of various colors according to the specified positions and color attributes. (b) Although the state-of-the-art layout-to-image method GLIGEN can generate ``donuts" according to the specified position in this example, their color attributes are not correct. We use boxes with ``Attr" to mark the wrong color attributes. (c) Our proposed MIGC can not only generate ``donuts" according to the position specified by the annotation but also ensure that the color attribute of each generated donut instance is correct.} 
  \vspace{-1mm}
    \label{fig:diff_MIG_pos}
  \end{figure}

\section{Construction Process of COCO-MIG Benchmark}

\noindent
\textbf{Overview.} COCO-MIG benchmark uses the layout of COCO-position benchmark~\cite{coco} and assigns a specific \textbf{color} attribute to each instance. COCO-MIG requires that each instance generated not only meet the position requirements but also meet the attribute (i.e., color) requirements.

\noindent
\textbf{Step 1: Sampling layouts from COCO.} We sample layouts from the COCO-position~\cite{coco}, filter out instances with side lengths less than 1/8 of the original image size, and further filter out those layouts with less than two instances. To test the model's ability to control quantity, we divided these layouts into five levels, $L_{2}$-$L_{6}$, based on the number of instances, where $L_i$ indicates that there are i instances in the target-generated image. A total of 160 layouts are sampled for each level. Notably, in the process of sampling layouts for level $L_i$, if the number of instances surpasses $i$, we selectively choose the initial $i$ instances with the largest area. Conversely, if the number of instances is less than $i$, a resampling procedure is employed. 

\noindent
\textbf{Step 2: Assigning color attribute to each instance.} On the basis of each sampled layout, we assign each instance a specific color from eight colors, i.e., red, yellow, green, blue, white, black, and brown. At the same time, we write the global prompt as 'a $<$attr1$>$ $<$obj1$>$, a $<$attr2$>$ $<$obj2$>$, ..., and a ...'.

\section{Difference between COCO-MIG and COCO-position}

\noindent
Fig.~\ref{fig:diff_MIG_pos} shows a specific example. In this example, COCO-MIG assigns a specific color to each ``donut" based on the COCO-position layout. Fig.~\ref{fig:diff_MIG_pos}(b) shows the results of the state-of-the-art layout-to-image method GLIGEN~\cite{gligen}. It can be seen that the results generated by GLIGEN meet the position requirements, so this will be judged as correctly generated in the COCO-positon benchmark. However, COCO-MIG not only requires the generated instances to meet position requirements but also attributes requirements. From this perspective, COCO-MIG will determine that the results generated by GLIGEN are incorrectly generated because there are ``donuts" of incorrectly generated color attributes. Finally, it can be seen that using our proposed MIGC guarantees that the position and attributes of each generated instance are correct.

\section{More MIG Results}
\begin{figure*}[htb]
    % \vspace{-2mm}
    \setlength{\abovecaptionskip}{0.cm}
    \setlength{\belowcaptionskip}{0.cm}
    \begin{center}
        \includegraphics[width=1.0\textwidth]{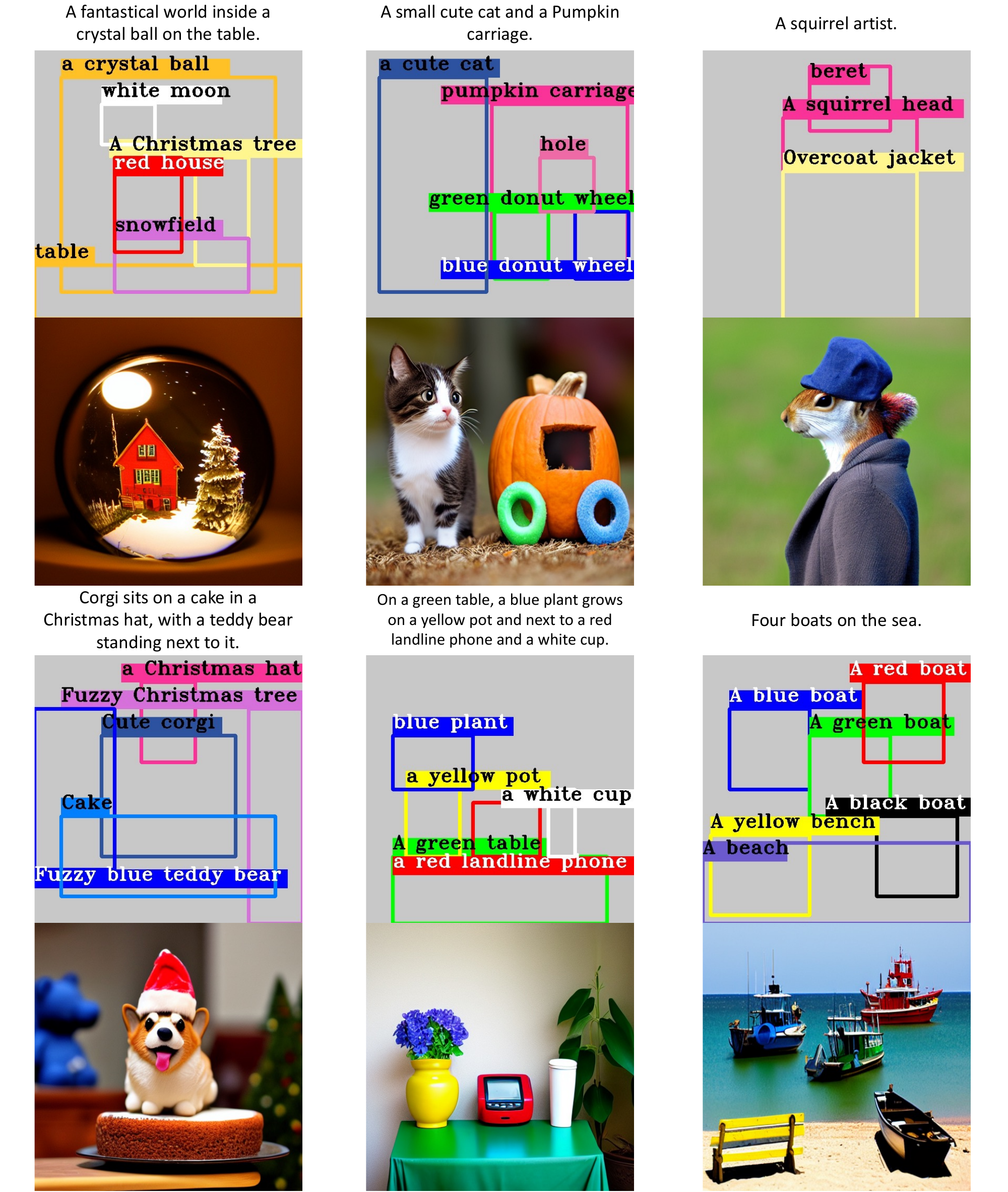}
    % \vspace{-1mm}
    \end{center}
   \caption{Multi-Instance Generation (MIG) with our MIGC. MIGC can generate images based on various complex layouts and ensure that the attributes of each instance are correct.} 
  \vspace{-1mm}
    \label{fig:more_MIG_1}
  \end{figure*}

\begin{figure*}[htb]
    % \vspace{-2mm}
    \setlength{\abovecaptionskip}{0.cm}
    \setlength{\belowcaptionskip}{0.cm}
    \begin{center}
        \includegraphics[width=1.0\textwidth]{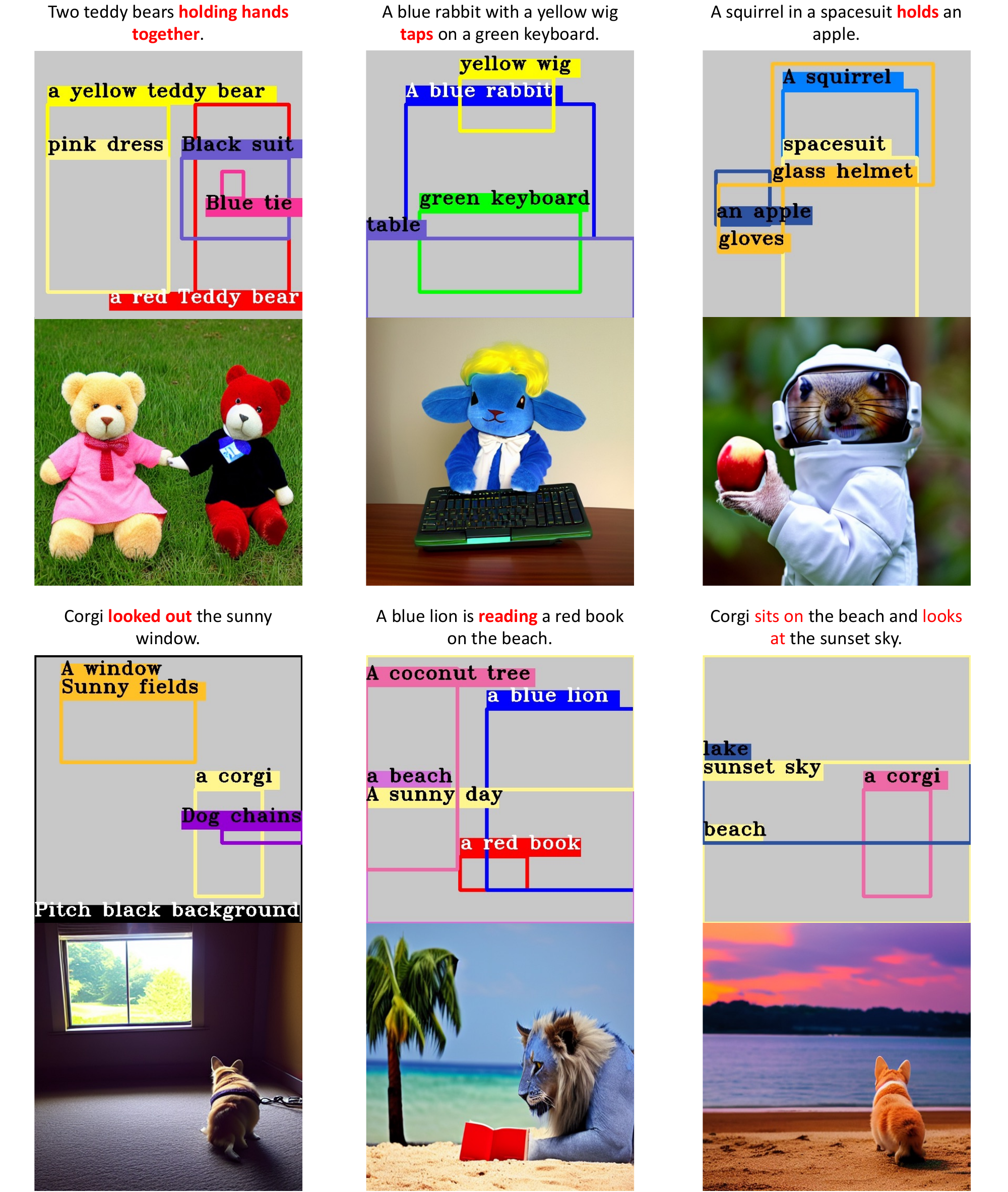}
    % \vspace{-1mm}
    \end{center}
   \caption{Multi-Instance Generation (MIG) with our MIGC. By specifying the relation between instances through the global prompt, MIGC can further control the interaction of instances.} 
  \vspace{-1mm}
    \label{fig:more_MIG_2}
  \end{figure*}
Fig.~\ref{fig:more_MIG_1} and Fig.~\ref{fig:more_MIG_2} show more results obtained using MIGC for Multi-Instance Generation. Even with complex layouts and rich attribute descriptions, MIGC can ensure that each instance is generated at the correct position and has the correct attributes. At the same time, if the relationship between each instance is specified in the global prompt (e.g., action relationship), MIGC can further control the interaction between instances.

\section{More Qualitative Results on COCO-MIG}
\begin{figure*}[htb]
    % \vspace{-2mm}
    \setlength{\abovecaptionskip}{0.cm}
    \setlength{\belowcaptionskip}{0.cm}
    \begin{center}
        \includegraphics[width=1.0\textwidth]{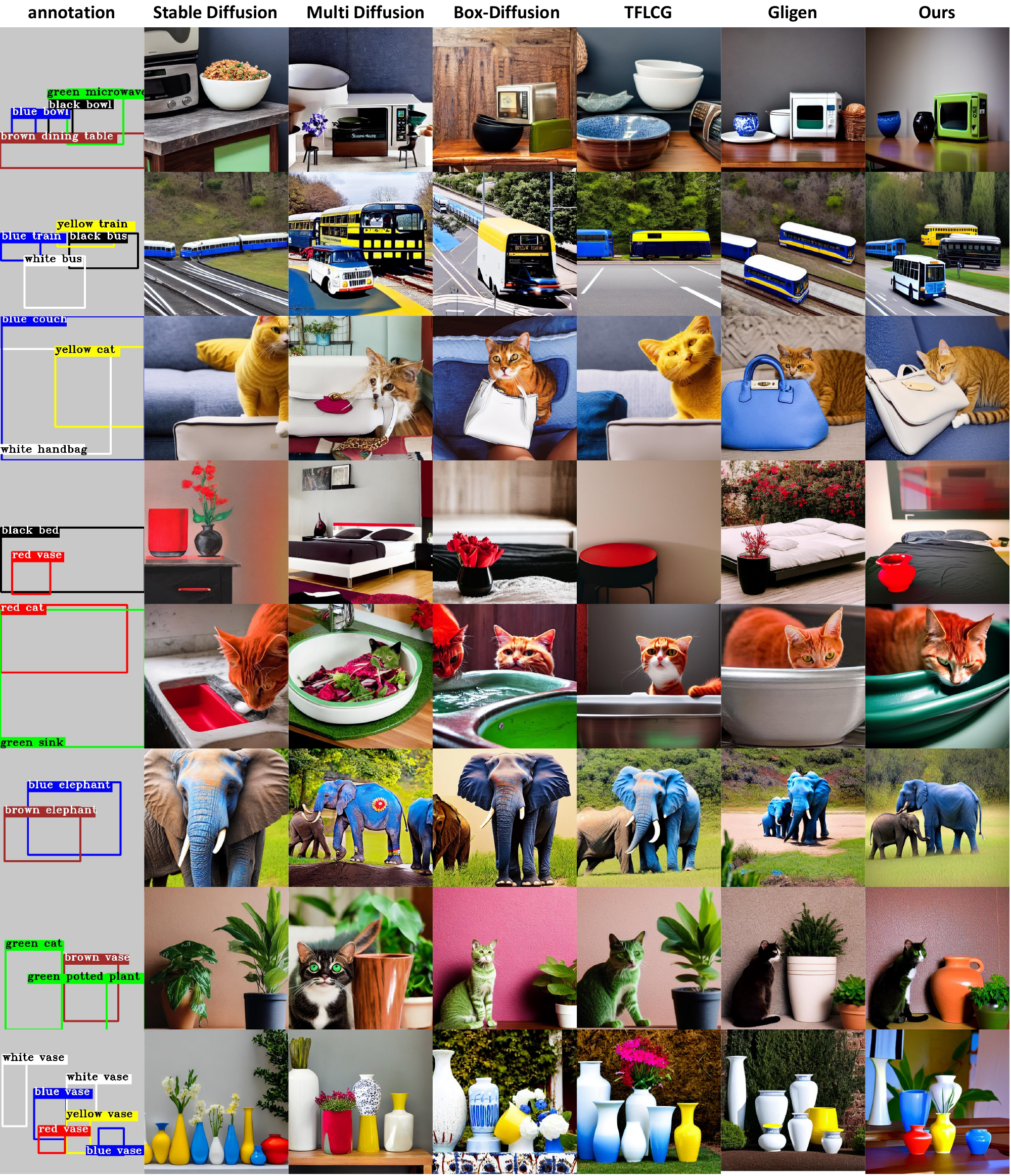}
    % \vspace{-1mm}
    \end{center}
   \caption{More qualitative results of the proposed COCO-MIG benchmark. The first column on the left shows the target layout and descriptions for each instance, and we use the corresponding colored boxes to describe the target color for each instance. The columns on the right display the results of baseline and MIGC, respectively. To better observe the details of the image, we did not add any extra annotations to the image.} 
  \vspace{-1mm}
    \label{fig:more_COCO_MIG}
  \end{figure*}

More qualitative results on our proposed COCO-MIG benchmark are shown in Fig.~\ref{fig:more_COCO_MIG}. Compared with previous state-of-the-art methods, our proposed MIGC approach can better control the position, attributes, and quantity simultaneously.

\section{Qualitative Results on DrawBench}

\begin{figure*}[tb]
    % \vspace{-2mm}
    \setlength{\abovecaptionskip}{0.cm}
    \setlength{\belowcaptionskip}{0.cm}
    \begin{center}
        \includegraphics[width=1.0\textwidth]{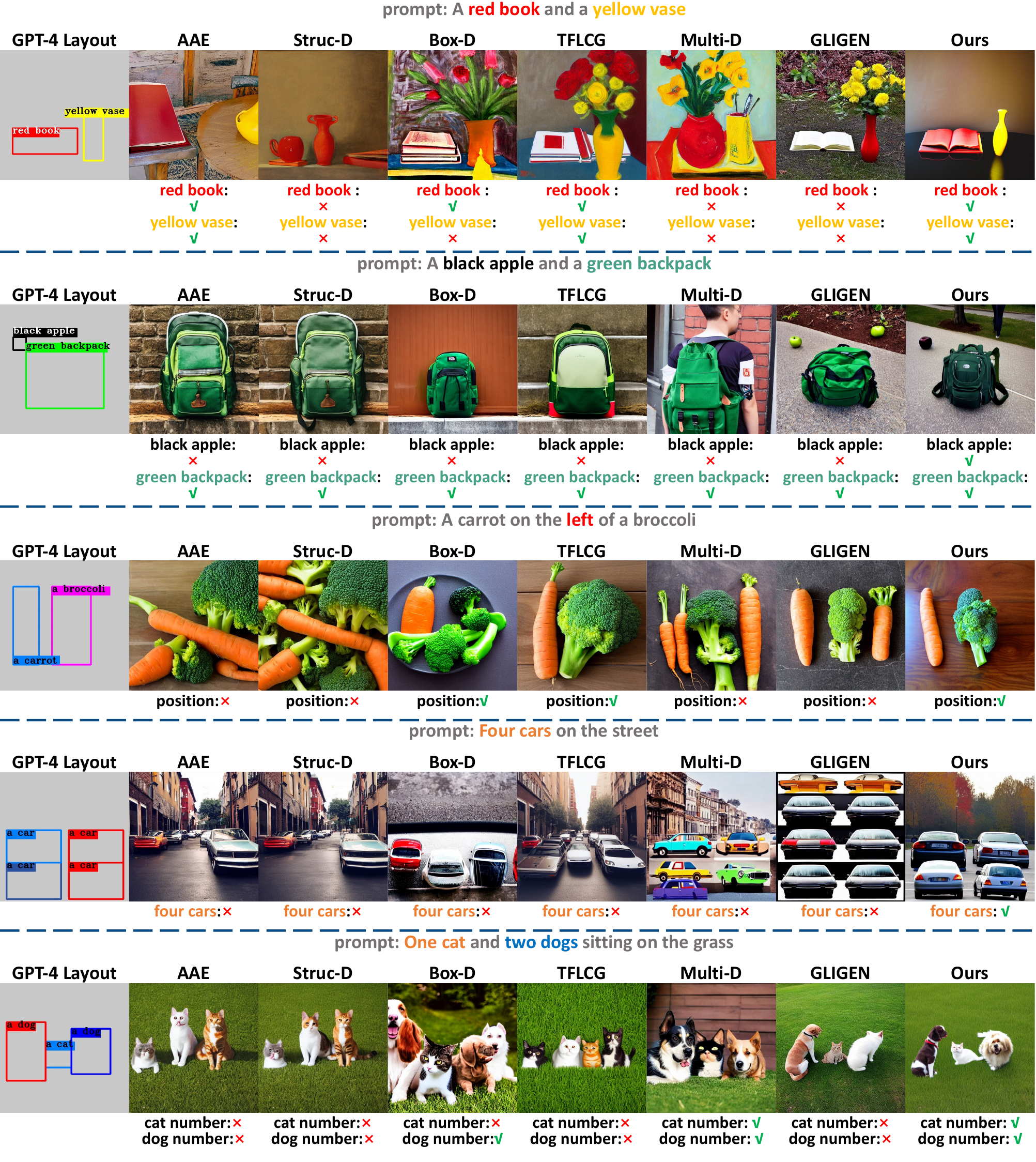}
    % \vspace{-1mm}
    \end{center}
   \caption{Qualitative comparison of DrawBench. The first and second rows show the results in the ``COLOR" subtask, in which MIGC achieves precise attribute control while obvious attribute leakage problems appear in other  state-of-the-art methods. The third row shows that MIGC can achieve precise position control. The fourth and fifth rows show that MIGC can achieve precise quantity control, especially in the case where ``cat" and ``dog" exist at the same time, in which MIGC avoids the mutual influence of cat and dog semantics since  MIGC can achieve good attribute control. } 
  \vspace{-1mm}
    \label{fig:drawbench_qualitative}
  \end{figure*}

\noindent
\textbf{Implementation details.}
DrawBench~\cite{imagen} is a challenging T2I benchmark. On this benchmark, we compare our proposed MIGC with state-of-the-art text-to-image (i.e., AAE~\cite{aae}, Struc-D~\cite{structurediff}) and layout-to-image methods (i.e., Box-D~\cite{boxdiff}, TFLCG~\cite{tflcg}, Multi-D~\cite{multidiffusion}, GLIGEN~\cite{gligen}). For the text-to-image methods, we directly input the DrawBench's prompt into the pipeline. For the layout-to-image methods and our proposed MIGC, we first use GPT-4~\cite{gpt4,layoutgpt} to generate the layout and then input it into the network, forming a two-stage text-to-image pipeline.

\noindent
\textbf{Qualitative results.}
Fig.~\ref{fig:drawbench_qualitative} shows the qualitative comparison on DrawBench. The first row shows that obvious attribute leakage problems (i.e., confusion between the ``yellow" and ``red" attributes) occur in the results of previous state-of-the-art methods, while our MIGC can control the attributes of each instance very precisely. The second row shows that previous state-of-the-art methods cannot correctly generate a ``black apple," which is counterfactual, while our MIGC can achieve good generation. The third row indicates that MIGC can control the position more accurately than the previous state-of-the-art method and can effectively solve the problem of extra generation (e.g., both Multi-D and GLIGEN have the phenomenon of excessive generation of ``carrots"), mainly due to the inhibition loss used in MIGC. The fourth row shows that MIGC can achieve accurate quantity control while other methods generate wrong quantities. The results in the fifth row show that when it comes to quantity control of multiple categories, stronger attribute control (e.g., it can avoid cat attributes from leaking to the dogs' region) makes MIGC achieve more accurate quantity control.

%-------------------------------------------------------------------

\begin{figure}[tb]
    % \vspace{-2mm}
    \setlength{\abovecaptionskip}{0.cm}
    \setlength{\belowcaptionskip}{0.cm}
    \begin{center}
        \includegraphics[width=0.5\textwidth]{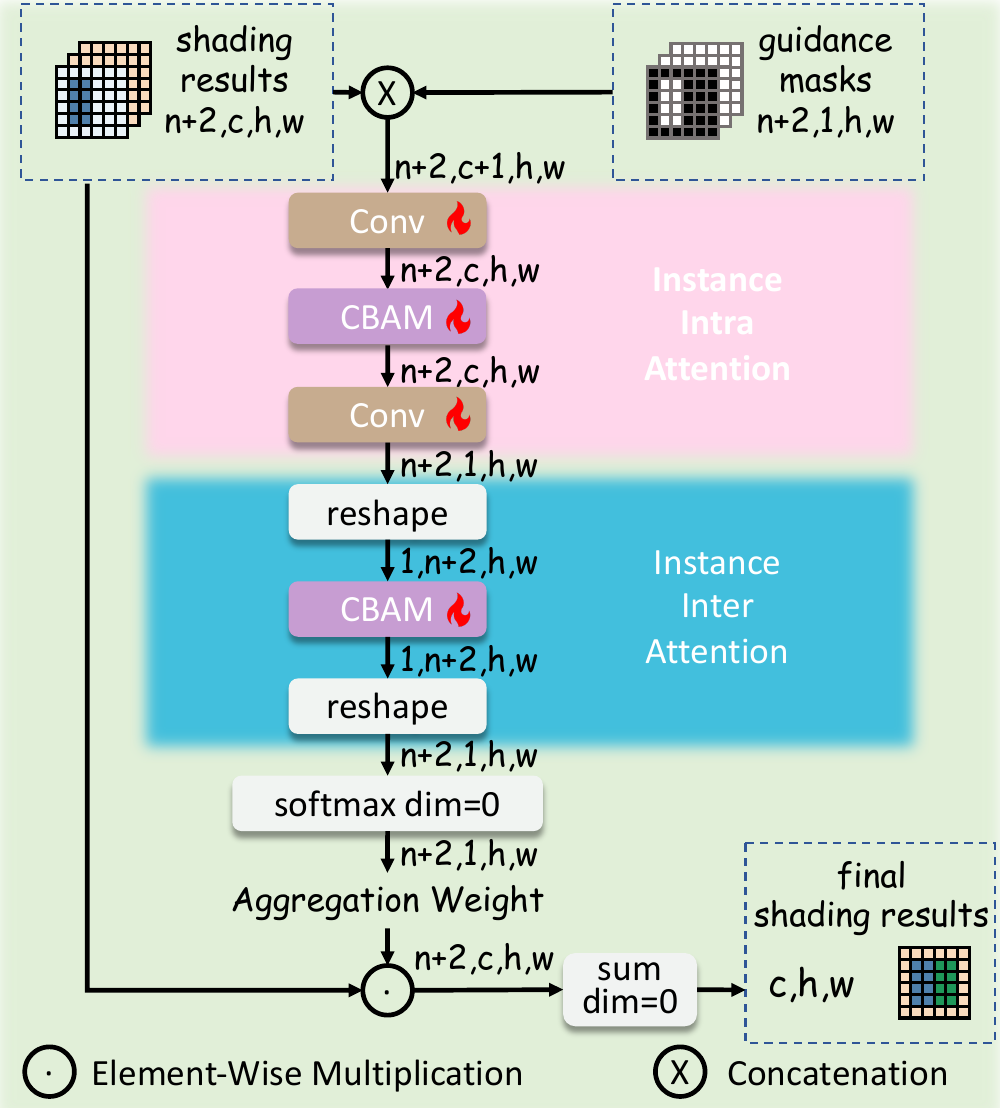}
    % \vspace{-1mm}
    \end{center}
   \caption{Architecture details of Shading Aggregation Controller.} 
  \vspace{-1mm}
    \label{fig:fig_SAC_details}
  \end{figure}

\begin{figure}[tb]
    % \vspace{-2mm}
    \setlength{\abovecaptionskip}{0.cm}
    \setlength{\belowcaptionskip}{0.cm}
    \begin{center}
        \includegraphics[width=0.5\textwidth]{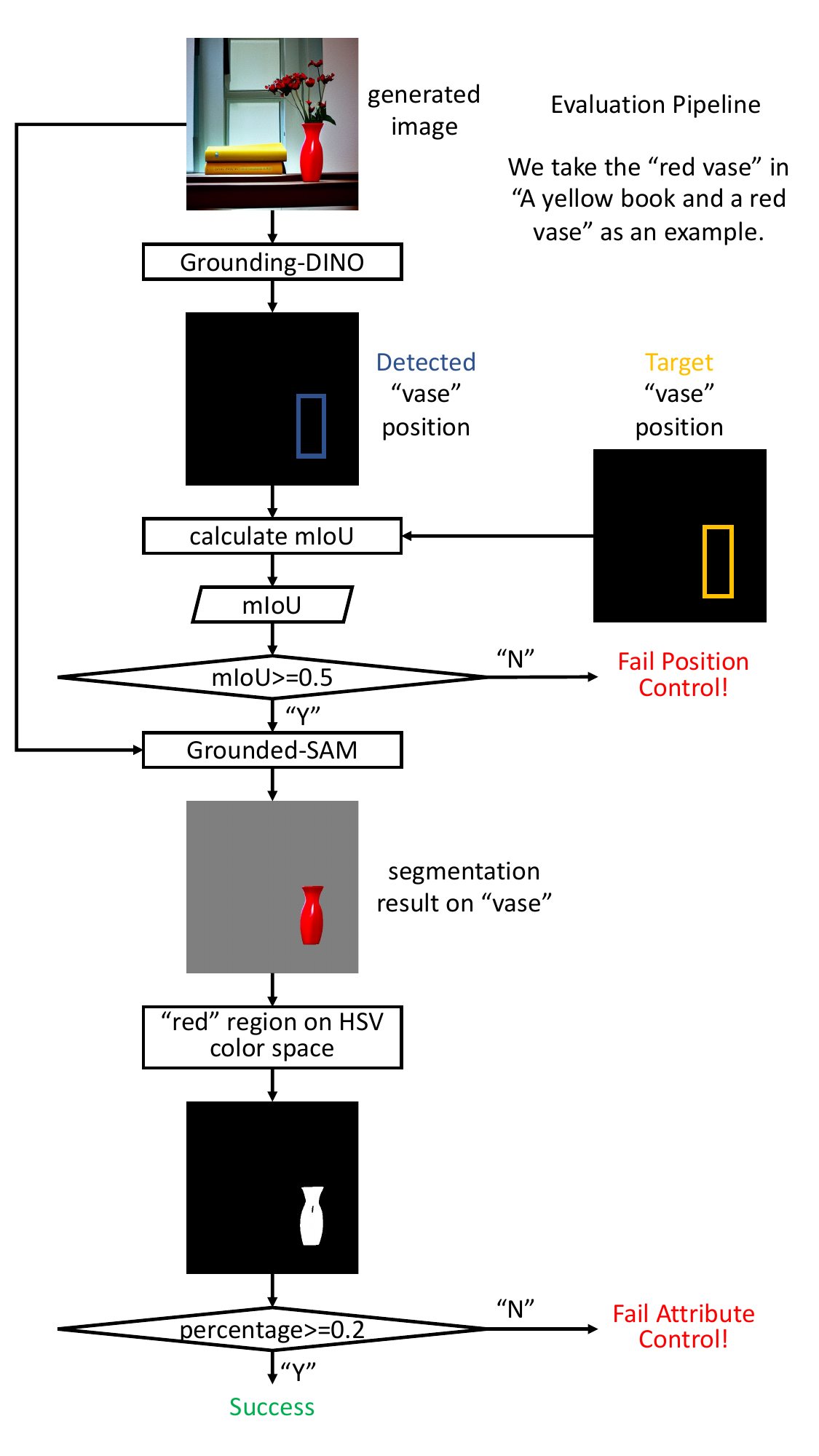}
    % \vspace{-1mm}
    \end{center}
   \caption{Evaluation Pipeline.} 
  \vspace{-1mm}
    \label{fig:eval_pipe}
  \end{figure}
\noindent

\begin{figure*}[tb]
    % \vspace{-2mm}
    \setlength{\abovecaptionskip}{0.cm}
    \setlength{\belowcaptionskip}{0.cm}
    \begin{center}
        \includegraphics[width=1.0\textwidth]{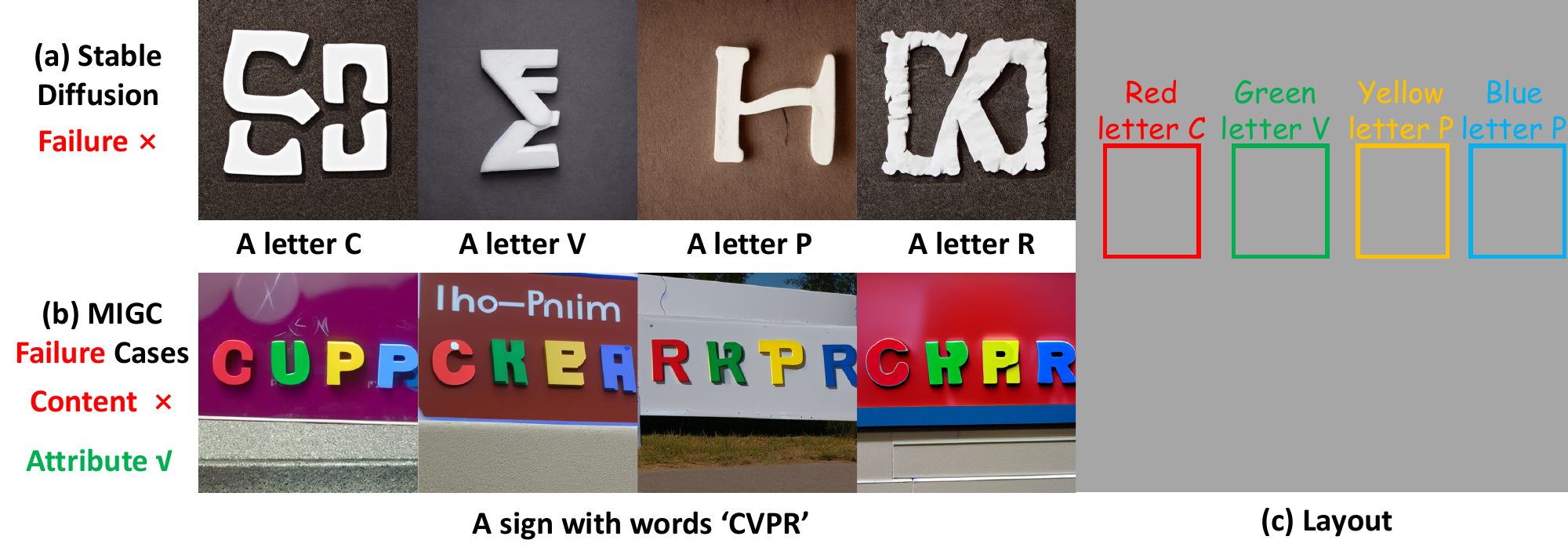}
    % \vspace{-1mm}
    \end{center}
   \caption{Limitation. (a) shows that stable diffusion struggles to generate individual letters `C,' `V,' `P,' and `R.' (b) shows the failure cases of MIGC. (c) shows the layout information of the failure case. If stable diffusion struggles to generate a specific instance, our MIGC will also encounter difficulties when generating this instance or its combination with other instances.} 
  \vspace{-1mm}
    \label{fig:limitation}
  \end{figure*}

\section{Details of Shading Aggregation Controller}
\noindent
\textbf{Overview.} Illustrated in Fig.~\ref{fig:fig_SAC_details}, after obtaining the shading results $\mathbb{R}_s=\{\mathbf{R}_{s}^{1},\ldots,\mathbf{R}^{N}_{s}, \mathbf{R}^{bg},\mathbf{R}_{LA}\}\in \mathbb{R}^{(N+2,C,H,W)}$ and guidance masks $\mathbb{M}=\{\mathbf{M}^{1},\ldots,\mathbf{M}^{N}, \mathbf{M}^{bg},\mathbf{M}_{LA}\}\in \mathbb{R}^{(N+2,1,H,W)}$, the Shading Aggregation Controller (SAC) sequentially performs instance \textbf{intra-attention} and \textbf{inter-attention} to dynamically aggregate shading results, and aggregation weights summing to 1 are assigned to shading results on each spacial pixel through the softmax function, resulting in the final shading $\mathbf{R}_{final} \in \mathbb{R}^{(H,W,C)}$.

\noindent
\textbf{Instance Intra-Attention.} As shown in Fig.~\ref{fig:fig_SAC_details}, after SAC concats the shading results and guidance masks in the channel dimension, it will perform instance intra-attention through a stack of Conv-CBAM-Conv layers, in which Conv layers are mainly used to change the channel number, and the CBAM~\cite{cbam} sequentially performs channel-wise and spatial-wise attention in each instance's feature map.

\noindent
\textbf{Instance Inter-Attention.} As shown in Fig.~\ref{fig:fig_SAC_details}, after instance intra-attention, the SAC further performs instance inter-attention, in which the SAC reshapes the features to change the dimension order of the feature map and then uses the CBAM to perform instance-wise attention.

%-------------------------------------------------------------------

%-------------------------------------------------------------------
\section{Details of Evaluation Pipeline}
\textbf{Overview.}
The flowchart in Fig.~\ref{fig:eval_pipe} shows the details of the evaluation pipeline, and we will introduce it by telling how to check whether  ``a red vase" is accurately generated.

\noindent
\textbf{Position Evaluation.}
First, we input the generated image into Grounding-DINO~\cite{gdino} to detect the bounding box of the ``vase" and then calculate the IoU with the target layout's bounding box. If the IoU$\geq$0.5, this "vase" is determined to be \textit{Position Correctly Generated}. Note that if multiple bounding boxes are detected in the generated image, we will select the one closest to the target layout's bounding box to calculate IoU.

\noindent
\textbf{Attribute Evaluation. }
After checking that the ``a red vase" is \textit{Position Correctly Generated}, we will further check whether its attribute (i.e., ``red" color) is generated accurately. Specifically, we will use Grounded-SAM~\cite{sam} to segment the ``vase" region in the generated image and mark its area as M. Then we will calculate the area in M that meets the ``red" requirement on the HSV color space and mark it as O. If the percentage O/M $\geq$ 0.2, we can consider that this ``red vase" has the correct attribute and mark it as \textit{Fully Correctly Generated}.

\noindent
\textbf{Evaluation for Different Benchmarks.}
Benchmarks requiring both attribute and position control, such as our proposed COCO-MIG, require each instance to be \textit{Fully Correctly Generated}. Benchmarks requiring only position control, such as COCO-positon, require each instance to be \textit{Position Correctly Generated}.

\section{Manual Evaluation on DrawBench}

We also perform a manual evaluation on DrawBench~\cite{imagen} to check whether the generated images adhere to the input text description in color, position, and count dimensions. Specifically, ten people will participate in the evaluation, and each generated image will be judged as "correctly generated" or "wrongly generated." We show the average accuracy calculated based on the evaluation results of ten people.

Different from automated evaluation, which strictly considers the mIoU to determine whether the local generation is successful or not, manual evaluation mainly checks whether the generated image satisfies the text description globally.

\begin{figure*}[htb]
    % \vspace{-2mm}
    \setlength{\abovecaptionskip}{0.cm}
    \setlength{\belowcaptionskip}{0.cm}
    \begin{center}
        \includegraphics[width=1.0\textwidth]{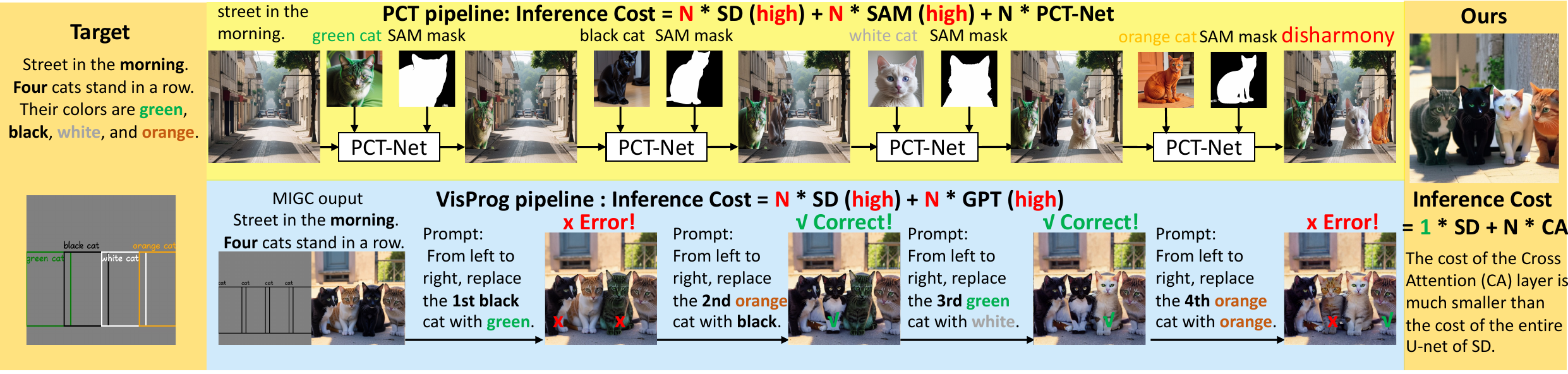}
    % \vspace{-1mm}
    \end{center}
   \caption{More baseline results.  Please zoom in for the best view.} 
  \vspace{-1mm}
    \label{fig:baseline_pipeline}
  \end{figure*}

\section{More Implementation Details}
\noindent
\textbf{Training.} We only deploy MIGC on the mid-layers (i.e., $8\times{8}$) and the lowest-resolution decoder layers of UNet(i.e., $16\times{16}$), which greatly determine the generated image's layout and semantic information~\cite{dragondiffusion,tflcg}. In other Cross-Attention layers, we use the global prompt for global shading. We use COCO 2014 \cite{coco} to train MIGC. To get the instance descriptions and their bounding boxes, we use stanza \cite{stanza} to split the global prompt and detect the instances with the Grounding-DINO\cite{gdino} model. To put the data in the same batch, we fix the number of instances to 6 during training, i.e., if data contains more than $6$ instances, 6 of them will be randomly selected. If data contains less than $6$ instances, we complete it with null text and coordinates [0.0, 0.0, 0.0, 0.0]. We train our MIGC based on the pre-trained stable diffusion v1.4. We use AdamW \cite{adam} optimizer with a constant learning rate of $1e^{-4}$, and train the model for 300 epochs with batch size 320, which requires 15 hours on 40 V100 GPUs with 16GB VRAM each. 

\noindent
\textbf{Inference.} We use EulerDiscreteScheduler~\cite{elud} with 50 sample steps and use our MIGC in the first 25 steps. We select the CFG scale\cite{Ho2022ClassifierFreeDG} as 7.5. As shown in Fig.~\ref{fig:fig_SAC_details}, the channel number of the second CBAM layer (i.e., CBAM in Instance Inter Attention) in the Shading Aggregation Controller is related to the number of input instances. In order to allow our MIGC to handle different numbers of instances, we set the channel number of the second CBAM layer as $max\_num+2$ (e.g., our default setting is 28+2, which can satisfy almost all practical applications). In actual inference, we assume that the number of instances to be processed is $n_{infer} \leq max\_num$, and we get $f_{intra}\in\mathbb{R}^{(1,n_{infer}+2,h,w)}$ through Instance Intra Attention. Next, we need to pad the number of channels of $f_{intra}$ to $max\_num+2$. Specifically, we enter an all-0 features $f_{zero}\in{\mathbb{R}^{1,c,h,w}}$ (i.e., this is consistent with the shading result of null text during the training) into Instance Intra Attention to get $f_{padding}=InstanceIntraAttenion(f_{zero}),f_{padding}\in{\mathbb{R}^{1,1,h,w}}$, and we use $f=concat([f_{intra}]+ [f_{padding}]*(max\_num-n_{infer}), dim=1),f\in{\mathbb{R}^{1,max\_num+2,h,w}}$ as the input of the CBAM layer in Instance Inter Attention. In order to allow the network to notice the later-ordered shading instances during actual inference, we will randomly shuffle the above $f_{padding}$ and $f_{intra}$ during training, while the shading background and shading template will not participate in the above shuffle process. At the same time, we can observe that since Instance Intra Attention has eliminated the larger number of original feature channels C (e.g., 1280), the computational complexity will be very low even when processing a larger number of shading instances in Instance Inter Attention.

\section{More Baselines}
Based on the divide-and-conquer idea, we also designed two other baselines. The qualitative comparisons are shown in Fig.\ref{fig:baseline_pipeline}

\noindent
\textbf{1)PCT-Net Pipeline.} As shown in the first row of Fig.\ref{fig:baseline_pipeline}, we first independently generate each individual instance and the background. Then, we use PCT-Net~\cite{pctnet}, a state-of-the-art image fusion network, to merge all instances with the background.
Using PCT-Net to fuse the pre-generated images ensures the correctness of attributes, which verifies the effectiveness of our divide-and-conquer idea.  
However, this pipeline incurs significant inference time costs, and the generated images may lack harmony.

\noindent
\textbf{2)Visual Programming Pipeline.} 
As shown in the second row of Fig.\ref{fig:baseline_pipeline},
Visual Programming~\cite{visprog} utilizes the GPT model to parse user input commands and generate a series of predefined operations, thereby achieving functionalities such as image editing. Here, we employ this method to sequentially perform editing on each object in the pre-generated images from left to right, aiming to correct the attributes of each object as much as possible. This pipeline is capable of correcting erroneously generated attributes. However, it uses text to locate and edit the instance, which lacks precise positioning capabilities and faces challenges in deployment to real scenes with complex layouts. For example, the 1st and 4th steps locate and affect the incorrect cat. In addition, these methods utilize GPT to coordinate image generation, making large-scale generation expensive and challenging.

\section{Limitation}

Inspired by the idea of divide and conquer, MIGC maximizes the use of the powerful Single-Instance Generation capability of pre-trained stable diffusion and extends it to MIG tasks. However, for a specific instance that stable diffusion cannot generate well, our MIGC will also encounter difficulties when generating this instance or its combination with other instances. As Fig.\ref{fig:limitation}(a) shows, stable diffusion has difficulty generating individual letters accurately. Therefore, when using MIGC to generate the words `CVPR' in the layout of Fig.\ref{fig:limitation}(c), we see that although MIGC correctly controls the color attribute of each letter, the content of the actual letters is wrong, causing the entire sample to fail, as shown in Fig.\ref{fig:limitation}(b).

\end{document}